%% file: arxiv20.tex
\documentclass[sigplan,10pt,screen]{acmart}

\acmConference[]{\textsf{pe-graph2seq}}{February 2020}{ArXiV}
\acmYear{2020}
\acmISBN{} 
\acmDOI{} 
\startPage{1}

\setcopyright{none}

\usepackage{booktabs} 

\usepackage{courier}            
\usepackage[scaled]{helvet} 
\usepackage{url}                  
\usepackage{paralist}
\usepackage{enumitem}      

\usepackage{breakurl}
\usepackage{amsfonts}
\usepackage{subfig}
\usepackage{subfloat}
\usepackage{stfloats}
\usepackage{clrscode}
\usepackage{xcolor}
\usepackage{listings}
\usepackage{tikz}

\usepackage[linesnumbered,ruled]{algorithm2e}

\usepackage{graphicx} 	
\usepackage{mathtools}	

\usepackage{multirow}	

\usepackage{amsmath}
\usepackage{amsthm}

\usepackage{bm}
\usepackage{xspace}

\input{macros}

\newcommand{\ignore}[1]{}

\definecolor{mygreen}{rgb}{0,0.6,0}
\definecolor{mygray}{rgb}{0.5,0.5,0.5}
\definecolor{mymauve}{rgb}{0.58,0,0.82}

\usepackage{caption}
\usepackage{rotating}

\usepackage{cancel}

\usepackage{balance}

\usepackage{ragged2e}
\usepackage{array}
\newcolumntype{R}[1]{>{\RaggedLeft\arraybackslash}p{#1}}
\newcolumntype{P}[1]{>{\centering\arraybackslash}p{#1}}

\usepackage[section]{placeins}

\begin{document}


\title[Equivalence of Dataflow Graphs Using Graph-to-Sequence Model]{Equivalence of Dataflow Graphs via Rewrite Rules\\ Using a Graph-to-Sequence Neural Model}




\author{Steve Kommrusch}
 \affiliation{
   \institution{Colorado State University}            
   \country{USA}                    
 }
 \email{steveko@cs.colostate.edu}          
\author{Th{\'e}o Barollet}
 \affiliation{
   \institution{Inria}            
   \country{France}                    
 }
 \email{theo.barollet@inria.fr}          
\author{Louis-No{\"e}l Pouchet}
 \affiliation{
   \institution{Colorado State University}            
   \country{USA}                    
 }
 \email{pouchet@colostate.edu}          


\begin{abstract}
\input{abstract}
\end{abstract}



\maketitle

\section{Introduction}
\label{sec:introduction}
\input{introduction}

\section{Motivation and Overview}
\label{sec:motivation}

\input{motivation}

\section{Framework for Program Equivalence}
\label{sec:proglangdefs}
\input{proglangdefs}

\section{Samples Generation}
\label{sec:samplegen}
\input{samplegen}
\section{Deep Neural Networks for Program Equivalence}
\label{sec:progequivdnn}
\input{progequivdnn}

\section{Experimental Results}
\label{sec:expresults}
\input{expresults}

\section{Related Work}
\label{sec:related}
\input{related}

\section{Conclusion}
\label{sec:conclusion}
\input{conclusion}

\begin{acks}                            
This work was supported in part by the U.S. National Science Foundation award CCF-1750399.
\end{acks}

\balance
\bibliography{lnp,arxiv20,bibs/refs,bibs/gabriel,bibs/ierefs,bibs/ics15}

\newpage

\appendix
\section{Appendix}
\input{appendix}

\end{document}

%% file: abstract.tex

In this work we target the problem of provably computing the equivalence between two programs represented as dataflow graphs. To this end, we formalize the problem of equivalence between two programs as finding a set of semantics-preserving rewrite rules from one into the other, such that after the rewrite the two programs are structurally identical, and therefore trivially equivalent. We then develop the first graph-to-sequence neural network system for program equivalence, trained to produce such rewrite sequences from a carefully crafted automatic example generation algorithm. We extensively evaluate our system on a rich multi-type linear algebra expression language, using arbitrary combinations of 100+ graph-rewriting axioms of equivalence. Our system outputs via inference a correct rewrite sequence for 96\% of the 10,000 program pairs isolated for testing, using 30-term programs. And in all cases, the validity of the sequence produced and therefore the provable assertion of program equivalence is computable, in negligible time.

%% file: introduction.tex


The problem of program equivalence is summarized as determining whether two programs would always produce the same output for all possible inputs, and is a central problem in computing \cite{kaplan1969regular,godlin2008inference,verdoolaege2009equivalence}. The problem ranges from undecidable, e.g. \cite{goldblatt2012well}, up to trivial in cases of testing the equivalence of a program with itself.


We claim the problem of program equivalence cannot be efficiently
mechanized by using a stochastic process to determine the equivalence
between two program regions. Precisely, obtaining a binary answer
yes/no to equivalence with a certain probability of confidence does
not lead to a provable conclusion on equivalence \cite{Xu17}. This is a major
limitation to the deployment of machine learning techniques for
program equivalence. Such an approach might prove useful for
e.g. filtering, to focus via another process on only a subset of
likely equivalent programs; but it is not a suitable approach for 
provably correct automated program equivalence checking as is typically developed, e.g., \cite{mansky2010framework,alias2004recognition,karfa2013verification}.


To overcome the fundamental stochastic nature of neural
networks, we use a very different approach to the problem of
machine learning for program equivalence: instead of making the
network produce a binary answer to the question of equivalence, \emph{we make the network produce a sequence of
  rewrite terms that make one program strictly equal to the other, if
  the input programs are equivalent}.
This way, the output of the
network can be deterministically checked in negligible time. We represent programs as graphs, and successively apply the axiom-based graph rewrites produced by the
network on one of the input programs, then ensure the resulting graph is identical
to the other input graph via a simple simultaneous depth-first visit.

Our neural network approach allows for
deterministically proving equivalence, entirely avoids false
positives, and quickly invalidates incorrect answers produced by the network
(no deterministic answer is provided in this case). In a
nutshell, we develop the first graph-to-sequence neural network system to
accelerate the search in the space of possible combinations of
transformation rules (i.e., axioms of equivalence in the input
language) to make two programs/graphs structurally identical without violating their original semantics.
We make the following contributions:
\begin{itemize}[noitemsep,topsep=0pt,wide=0pt]
\item We propose a machine learning system for program equivalence which ensures
correctness for all non-equivalent programs input, and a deterministically checkable output for equivalent programs.
  
\item We introduce \texttt{pe-graph2seq}, the first graph-to-seq-uence neural network system targeting program equivalence to the best of our knowledge. We provide the first implementation of such graph-to-sequence systems in the popular OpenNMT-py framework \cite{opennmt}.

\item We present a complete implementation of our system operating on a rich language for multi-type linear algebra expressions. Our system provides a correct rewrite rule sequence between two equivalent programs for 96\% of the 10,000 test cases, for a typical inference time of 16ms per pair of programs. The correctness of the rewrite rule is deterministically checkable in all cases in negligible time.
  
\end{itemize}

The rest of the paper is organized as follows. Sec.~\ref{sec:motivation} outlines the program equivalence problem we address, and motivates our proposed approach. Sec.~\ref{sec:proglangdefs} formally defines the type of program representation and axioms of equivalence we manipulate in this work, and formalizes the equivalence problem addressed. Automatic sample generation is discussed in Sec.~\ref{sec:samplegen} before Sec.~\ref{sec:progequivdnn} which introduces \texttt{pe-graph2seq}, its overall design principles and key components. A complete experimental evaluation of our system is  detailed in Sec.~\ref{sec:expresults}. Related work is presented in Sec.~\ref{sec:related} before concluding.

%% file: motivation.tex
\input{figexample1.tex}

\paragraph*{Input program representation}
Figs.~\ref{fig:treeexamples:1}-\ref{fig:treeexamples:4} show four examples of simple computations. For example, Fig.~\ref{fig:treeexamples:1} models the expression $a(1b+1c)$, one can imagine it to be the result of $a(db+dc)$ after e.g. constant-propagation of $1$ to $d$. In the following we call these equivalently programs, sentences from a language, and graphs, the reader needs to be ready to jump between these equivalent representations. They are defined by a single root, have nodes which can be operations consuming the value of their immediate predecessor or terminal/input values, and a node produces a value that can be used by its immediate successors. In essence this is a classical dataflow representation of the computation \cite{buck1993scheduling}, and what our system uses as input program representation. 


\paragraph*{Rewrite rules as axioms of equivalence}
Consider the programs in Fig.~\ref{fig:treeexamples:1} versus Fig.~\ref{fig:treeexamples:2}. The multiplication of a value by $1$ does not change the value, if we rely on an axiom of equivalence stating $1 * x = x,~\forall x \in \mathbb{N}$. This axiom specifies a strict criterion of application: the node must be of type $\mathbb{N}$, the expression pattern must be $1*x$; and a strict rewrite rule: replace a sub-graph $1*x$ for any $x$ by the graph $x$. In other words, replacing $1*b$ by $b$ in Fig.~\ref{fig:treeexamples:1} is a semantics-preserving rewrite, from the axiom of equivalence. In this work we view the problem of program equivalence as finding a sequence of semantics-preserving rewrites, each from a precisely defined axiom of equivalence, that rewrites one program into the other. If one program can be rewritten by a sequence of individually-correct semantics-preserving transformations into another one, then not only are they equivalent under the set of axioms used, but the sequence forms the constructive and verifiable proof of equivalence.




\paragraph*{An example} In this work we illustrate and experimentally evaluate our system using a rich linear algebra expression language because it exposes clearly (and intuitively) the various key concepts that must be handled: (1) operating on dataflow graphs as input, supporting transformations that can (2) delete or (3) create new nodes in the graph, and transformations that (4) manipulate entire subtrees. We also wanted a language with (5) multiple variable types, e.g. scalars, vectors and matrices and (6) a large number of different operators with (7) distinct axioms applicable for each. All of these are captured in the language we experiment with, see Sec.~\ref{sec:proglangdefs} for its formal definition.

When applying the axiom $A1:1 * x = x,~\forall x \in \mathbb{N}$ on the program $P$ in Fig.~\ref{fig:treeexamples:1} for its node $b$, we obtain an equivalent and yet syntactically different program,  we have $P \equiv A1(b,P)$. Applying the same axiom $A1$ on $c$ in the resulting program leads to program $P'$ in Fig.~\ref{fig:treeexamples:2}, and $P'\equiv P \equiv A1(c,A1(b,P))$. Precisely, in graph terms, Fig.~\ref{fig:treeexamples:2} is the result of a sequence of two semantics-preserving node deletion operations, as defined in the axiom.

Consider now the axiom $A2:x * (y+z) = x*y+x*z,~\forall x,y,z \in \mathbb{N}$. This is the standard distributivity axiom on natural arithmetic. In terms of graph transformations, this is a complex rewrite: a new node is created ($*$), one node is moved ($+$ to the root), and edges are significantly modified. When this complex, but semantics-preserving, rewrite is applied to Fig.~\ref{fig:treeexamples:2}, we obtain Fig.~\ref{fig:treeexamples:3}, that is $P \equiv A2(*,A1(c,A1(b,P)))$.

Finally consider the axiom $A3:x+y = y+x,~\forall x,y \in \mathbb{N}$, the standard commutativity axiom for $+$. The graph transformation does not change the number of nodes nor edges, instead only alters two specific edges. Note that as the previous axioms, it also illustrates operations on sub-graphs: indeed $x$ and $y$ do not need to be input/terminal nodes, they can be any subgraph producing a value of the proper type. This is illustrated by applying on
Fig.~\ref{fig:treeexamples:3} to obtain Fig.~\ref{fig:treeexamples:4}, that is the computation $ac+ab$. We have 
$P \equiv A3(+,A2(*,A1(c,A1(b,P))))$, a verifiable proof of equivalence under our axioms between the programs $a(1b+1c)$ and $ac+ab$, which involved structural changes including node deletion, creation and edge modification. Note the bidirectional nature of the process: one can rewrite from $a(1b+1c)$ to $ac+ab$, or the converse using the same (but reverted) sequence. Note also the non-unicity of a sequence: by possibly many ways a program can be rewritten into another one, for example the sequence $P \equiv A3(+,A1(c,A1(b,A2(*,P))$ also correctly rewrites Fig.~\ref{fig:treeexamples:1} into Fig.~\ref{fig:treeexamples:4}. Conversely, a sequence may not exist: for example no sequence of the 3 above axioms allow to rewrite $a+b$ into $a*b$. We call these non-equivalent in our system, that is precisely if there is no sequence of axioms that can be applied to rewrite one program into the other.





\paragraph*{The need for a verifiable procedure} A key motivation of our work is to enable in a safe and provably correct way the use of machine learning for program equivalence. For full automation of the process, we focus on ensuring correctness in case an equivalence result is computed by the system. That is, our system by design answers only with a probability of confidence that the two programs are not equivalent, but \emph{it produces a verifiable procedure to assess equivalence} otherwise. We believe such an approach is key for a practical, automated deployment of neural networks for program equivalence: verifiably proving equivalence to ensure no false positive, while tolerating a moderate amount of false negative (i.e., missing that two programs were in fact equivalent).

Numerous practical applications of the kind of system we develop exist, even on the linear algebra language we demonstrate on: for example the automatic correction of exercises for students, where they typically need to prove equivalence between two formulas by successive application of other formulas/axioms. Languages like e.g. Matlab could use interactive checking of the equivalence between the expression being typed and the pre-existing library implementations (e.g., BLAS-based \cite{goto2008high}) to use instead accelerated implementations when possible in real-time. But we have designed and evaluated our system in a robust enough way to be applicable to a wide variety of languages and problems, as long as they can be cast in the framework in Sec.~\ref{sec:proglangdefs}. We discuss other uses cases and applications in Sec.~\ref{sec:expresults}.


\paragraph*{The space of equivalences} Intuitively, our approach to program equivalence is as follows. We can intellectually reason on a graph for equivalent programs where each node represents a distinct program in the language, and two nodes (i.e., two different programs) are connected by a directed edge iff the source node can be rewritten as the target node by the application of a single one of the pre-defined axioms for equivalence. The edge is labeled by the axiom used and the specific position in the source node's program to where it needs to be applied to obtain the program in the target node.
Then there will be one or more paths in this graph from the two nodes modeling the two input programs if they are equivalent (one can be rewritten into the other while preserving semantics); and no path if no such rewrite is possible, that is the programs would be not equivalent in our framework. Exposing a path between two nodes is sufficient to prove the equivalence of their associated programs.
%

This path is exactly a sequence of rewrite rules from one program to another. To test the correctness of an arbitrary sequence, i.e., verify if this path exists in the graph and assess equivalence if it does, one then needs to simply apply the proposed sequence to one of the input programs: verify at each step that the rewrite in the sequence is indeed applicable (by a simple check of the applicability of the axiom at this particular program point), and eventually ensure the rewritten program is identical to the other input one. This test can be computed in time mostly linear with the program size in our framework, and when successful it implements a constructive proof of equivalence between the two programs.

\paragraph*{Pathfinding equivalence proofs}
When formulating the program equivalence problem this way, we can then view its solution as learning how to build at least one feasible path between any two pairs of nodes in the above graph, when it can exist. We can see that by design, there is a lot of redundancy in this space: the same labeled path will occur between many different pairs of programs (e.g., those where only the variable symbols differ), and there are typically many paths between the same two (equivalent) programs. This creates opportunities for the system to learn program representation and path construction techniques more easily.

Our key contribution is the development of a deep learning framework that learns this procedure automatically. The neural network system we build is trained by randomly sampling this graph, with samples made of two nodes and a path between them when training on equivalent programs, and an empty path otherwise. We specifically learn a generalization of the problem of finding paths in this graph as follows. We represent input programs in a carefully-crafted normalized dataflow-like graph encoded as a gated graph neural network \cite{Scarselli09,Beck18}, to enable structural, size-tolerant reasoning by the network on the inputs. It is combined with a global attention-based mechanism and a memory-based LSTM \cite{Hochreiter97} decoder which can memorize graph changes for producing the rewrite sequence and enable path-size tolerant reasoning, while following the properties of the axioms for equivalence.

In a nutshell, we make the network learn a stochastic approximation of an iterative algorithm that would be able to construct a feasible path (when possible) between any two pairs of nodes in this equivalence graph, but trained simply by randomly sampling pairs of nodes and one carefully labeled path between them. This avoids entirely the need to craft smart exploration heuristics to make this path-finding problem feasible in practice. This is instead what we let the neural network learn automatically; and specifically why we implemented graph neural networks to solve this problem \cite{Scarselli09,Xu17}. We rely on the network to suggest a transformation path by inference, and then verify its validity in linear time.

\begin{figure*}
\includegraphics[width=15cm]{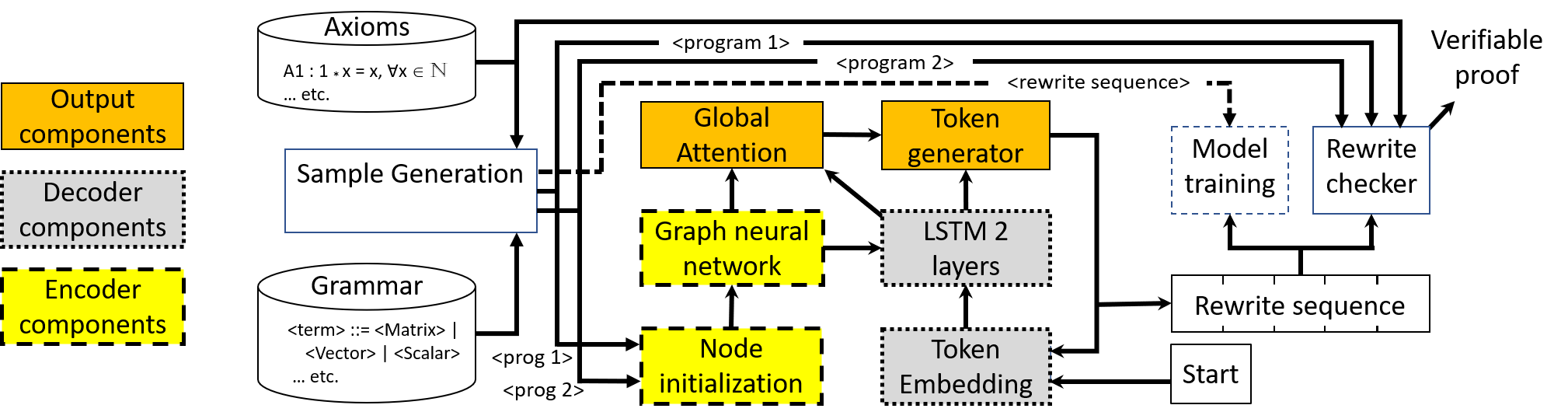}
\caption{\texttt{pe-graph2seq} System Overview}
\label{fig:Full}
\end{figure*}

\paragraph*{System overview} In order to implement our approach, we need a simple-enough grammar for a language, in which we enumerate randomly valid sentences, and a set of axioms of equivalence between two sentential forms expressible as semantics-preserving rewrite rules from one to the other. The system takes as input two programs (i.e., sentences accepted by the language), and produces an ordered sequence of axioms along with their position of application  (or node) that can be used to rewrite sequentially one input program into the other input program. This sequence is then checked for correctness using the axioms as reference. To train the system, we generate pairs of equivalent programs by iterating the axioms with random probability on one program, thereby generating both a path to equivalence and the target program. Random programs are generated so as to respect the grammar defined. The training set is then appropriately selected from these random samples, as detailed in Sec.~\ref{sec:expresults}.


When the system outputs that two programs are equivalent, as provable reasoning (the rewrite sequence) is always produced, no false positive can ever occur. When the system fails to demonstrate equivalence, no provable conclusion is produced regarding the input programs however: we are left with only a (high) probability of non-equivalence. 

A key of our approach is to introduce graph-to-sequence neural networks to quickly compute one or more possible rewrite sequences. The details of the network are covered in Sec.~\ref{sec:progequivdnn}. In a nutshell, the key principle is to combine a memory-based neural network approach, e.g., using Long-Short Term Memory (LSTM) \cite{Hochreiter97} neurons and a graph neural network design (which uses Gated Recurrent Units (GRUs) internally) \cite{Beck18} that matches our program graph representation. We use a sequence generation principle, using an attention mechanism to allow observation of program graph node information while generating the rewrite sequence. This enables the production of the rewrite sequence token-by-token, matching our axiom of equivalence design principle.

As presented in Sec.~\ref{sec:progequivdnn}, we designed an efficient embedding of the program equivalence problem into a graph neural network (\textsf{Node initialization}) to facilitate the network's ability to walk the program graphs and memorize structural changes being made by applying the axioms. To the best of our knowledge, this is the first graph-to-sequence system for program equivalence, which outputs verifiable proofs. Our system is fully implemented end-to-end in OpenNMT-py and is ready for artifact evaluation. As detailed in Sec.~\ref{sec:samplegen} we developed a very rigorous evaluation (test) set to ensure our system has developed some intelligence, which is fully confirmed in Sec.~\ref{sec:expresults}. As an extreme case, even if we would make a system that each time two programs are checked for equivalence, produces systematically all possible distinct rewrite sequences it has ever seen during training (up to 118k in our experiments), and we check all of these outputs for correctness and applicability for the input programs, this would still not exceed 60\% of correct test cases for equivalent programs tested. We report a steady 95\% or more using only a beam size of 10, that is asking the network to output only the 10 most likely rewrite sequences for the input programs.

%% file: figexample1.tex
\begin{figure*}
    \centering

    \centering
      \subfloat[][$a*(1* b+1* c)$]{
      \begin{minipage}[b]{0.25\textwidth}
        \label{fig:treeexamples:1}
        \centering
        \begin{tikzpicture} [level distance=1em, inner sep=1pt, minimum size=1.25em, edge from parent/.style={draw,latex-}]
            \node [circle, double, draw] {$*$}
            child [sibling distance=3.5em] {node [circle, draw] {$a$}
            child{edge from parent[draw=none] node [opacity=0] {}
            child  [sibling distance=2em]{edge from parent[draw=none] node [opacity=0] {}}
            child  [sibling distance=2em]{edge from parent[draw=none] node [opacity=0] {}}
            }
            child{edge from parent[draw=none] node [opacity=0] {}}
            }
            child [sibling distance=3.5em] {node [circle, draw] {$+$}
            child [sibling distance=3.5em, level distance=1em] {node [circle, draw] {$*$}
            child [sibling distance=2em, level distance=1.75em] {node [circle, draw] {$1$}}
            child [sibling distance=2em, level distance=1.75em] {node [circle, draw] {$b$}}
            }
            child [sibling distance=3.5em, level distance=1em] {node [circle, draw] {$*$}
            child [sibling distance=2em, level distance=1.75em] {node [circle, draw] {$1$}}
            child [sibling distance=2em, level distance=1.75em] {node [circle, draw] {$c$}}
            }
            };
        \end{tikzpicture}
        \end{minipage}
      }
      \unskip\ \hfill \vrule \hfill
      \subfloat[][$a*(b+c)$]{
      \begin{minipage}[b]{0.20\textwidth}
        \centering
        \begin{tikzpicture} [level distance=1.75em, inner sep=1pt, minimum size=1.25em, sibling distance=2em, edge from parent/.style={draw,latex-}]
            \node [circle, double, draw, align=center] {$*$}
                child {node [circle, draw] {$a$}
                    child{edge from parent[draw=none] node [opacity=0] {}}
                    child{edge from parent[draw=none] node [opacity=0] {}}
                }
                child {node [circle, draw] {$+$}
                    child{node [circle, draw] {$b$}}
                    child {node [circle, draw] {$c$}}
                };
        \end{tikzpicture}
        \label{fig:treeexamples:2}
        \end{minipage}
      }
    \unskip\ \hfill \vrule \hfill
    \subfloat[][$a * b + a * c$]{
      \begin{minipage}[b]{0.24\textwidth}
        \centering
        \begin{tikzpicture} [level distance=1.75em, inner sep=1pt, minimum size=1.25em, sibling distance=3.5em, edge from parent/.style={draw,latex-}]
            \node [circle, double, draw] {$+$}
                child {node [circle, draw] {$*$}
                    child [sibling distance=2em] {node [circle, draw] {$a$}}
                    child [sibling distance=2em] {node [circle, draw] {$b$}}
                }
                child {node [circle, draw] {$*$}
                    child [sibling distance=2em] {node [circle, draw] {$a$}}
                    child [sibling distance=2em] {node [circle, draw] {$c$}}
                };
        \end{tikzpicture}
        \label{fig:treeexamples:3}
        \end{minipage}
      }
    \unskip\ \hfill \vrule \hfill
    \subfloat[][$a * c + a * b$]{
      \begin{minipage}[b]{0.24\textwidth}
        \centering
        \begin{tikzpicture} [level distance=1.75em, inner sep=1pt, minimum size=1.25em, sibling distance=3.5em, edge from parent/.style={draw,latex-}]
            \node [circle, double, draw] {$+$}
                child {node [circle, draw] {$*$}
                    child [sibling distance=2em] {node [circle, draw] {$a$}}
                    child [sibling distance=2em] {node [circle, draw] {$c$}}
                }
                child {node [circle, draw] {$*$}
                    child [sibling distance=2em] {node [circle, draw] {$a$}}
                    child [sibling distance=2em] {node [circle, draw] {$b$}}
                };
        \end{tikzpicture}
        \label{fig:treeexamples:4}
      \end{minipage}
      }
    \caption{Examples of Computations}
    \label{fig:treeexamples}

\end{figure*}

%% file: proglangdefs.tex

We now present the formalism we use in this work to represent programs and their equivalences. We carefully co-designed this problem representation and the (graph) neural network approach to make the best use of machine learning via deep networks, as discussed in Sec.~\ref{sec:progequivdnn}. 


\subsection{Program Representation}

A key design aspect is to match the capability of the neural network to model the input as a walkable graph with the actual input program representation to be handled. We therefore model programs in a dataflow-like representation (i.e., a directed graph), using a single root/output node. In particular, we do not restrict to tree-like structures nor acyclic graphs, as briefly discussed in Sec.~\ref{sec:ruminations-on-extensions}. 

\begin{definition}[Program graph node]
  \label{def:proggraphnode}
A node $n \in N$ in the program graph models n-ary operations and input operands. A node produces a value which can be consumed by any of its immediate successors in the graph. When a node has no predecessor, it models an input value. The output value for the computation is produced by the unique root node $n_{root}$ of the graph, the only node without successor.
\end{definition}

\begin{definition}[Program graph directed edge]
  \label{def:proggraphedge}
  A directed edge $e_{n_1,n_2} : n_1 \rightarrow n_2$ with $n_1, n_2 \in N$ in the program graph connects the producer of a value ($n_1$) to a node consuming this value in the computation. 
\end{definition}

\begin{definition}[Program graph]
  \label{def:proggraph}
A program graph $G$ is a directed dataflow graph modeling the computation, made of nodes $n_i \in N$ and edges $e_{n_i,n_j} \in E$ as defined in Def.~\ref{def:proggraphnode} and Def.~\ref{def:proggraphedge}. That is,  $G = \langle n_{root}, N, E \rangle$. There is no dangling edge nor unconnected node in $G$. 
\end{definition}

\paragraph*{Language of linear algebra expressions} We developed a complex-enough language to evaluate carefully our work, that captures rich linear algebra expressions. Specifically, we support 3 types of data/variables in the program: scalars, vectors and matrices. We use the standard notation $a,\vec a,A$ for scalars, vectors and matrices.
We evaluate using different variable names for each of the 3 types above, along with their identity and absorbing elements.

We also model a rich set of operators, mixing different unary and binary operations for each type. Specifically, we support $*_s,+_s,-_s,/_s$ between scalar operands, and $+_v,-_v,*_v$ between vectors and $+_m,-_m,*_m$ for matrices. For $-,/$ we also support their unary version for all types, e.g. $^{-1_{s}}$ for unary scalar inversion and $-_{um}$ for unary matrix negation. For example $a^{-1_s}$ computes to $1/a$.
We also support multi-type operations, such as vector and matrix scaling by a scalar $*_{sv}, *_{sm}$. We support two specific unary matrix operations, transpose $^{t_m}$ and matrix inversion as $^{-1_m}$. Note every operator has a unique name in our language, driven by the type of its operand. This will facilitate the learning of the program embedding, avoiding the need to learn type propagation.

\paragraph*{Examples} Programs of the form $A (B C^t D) E^{-1}$, $\vec a + b\vec c^{-1}-0\vec e$, $(a+b)+(c(d/e))$, $(aA+bB)C^t$ etc. can be parsed trivially to our representation, one simply needs to be able to provide a unique name for each operand and operator type (possibly via some analysis, or simple language design principles), that is avoiding to overload the semantics of operators and operands. Note the semantics is never explicitly provided to our DNN approach, it is learned by examples. There will be no example of the form e.g. $a+A$, an invalid program in our language.

We believe a sensible approach is to develop a clean, regular grammar for the language to be handled, as implicitly these are concepts the DNN will need to learn. We did so, using a classical LL(1) grammar description of our linear algebra language. This is not a requirement of our approach, as one can arrive to the desired input program graph by any means necessary, but we believe making the reasoning on the language structure ``easy'' is an important design aspect.

\subsection{Axioms of Equivalence}

A central aspect of our approach is to view the problem of program
equivalence as finding a sequence of locally-correct rewrite rules
that each preserve the semantics, \emph{thereby making incremental reasoning possible}. We explicitly do not
consider non-semantics-preserving axioms. A rich structure of alternate but
equivalent ways to rewrite one program to another makes the problem
easier to sample and more amenable to machine learning. Semantics-preserving axioms enable incremental per-axiom reasoning, and enforce semantics preservation without overly complicated semantics analysis; while still manipulating a very
rich space of transformations. To illustrate this we specifically
design axioms that perform complex graph modifications, such as node
deletion or creation, subtree manipulation, multi-node graph changes,
	 etc. 


	 A graph pattern can be viewed as a pattern-matching rule on graphs and its precise applicability criteria. It can also be viewed as a sentential form of the language grammar, e.g. \texttt{ScalarVal PlusOp ScalarVal} is a pattern, if the grammar is well formed.


	 \begin{definition}[Graph pattern]
	 \label{def:graphpattern}
	 A graph pattern $P$ is an unambiguous structural description of a (sub-)graph $G_P$, which can be deterministically matched in any program graph $G$. We have $P = \langle G_P, M_n, M_e \rangle$ where for each node $n_i \in N^{G_P}$, $\{n_{match}\} = M_n(n_i)$ returns the set of node values $n_{match}$ accepted to match $n_i$ on a graph $G$. For $n_i,n_j \in N^{G_P}$, $e_i = M_e(n_i, n_j)$ returns the set of edges between $M(n_i)$ and $M(n_j)$ to be matched in $G$. A pattern $G_P$ is matched in $G$ if (a) $\forall n_i \in G_p,~ \exists~ n_m = M(n_i) \in N^G$; (b) $\forall e_i \in E^{G_P}, \exists~ e_{M_n(n_i),M_n(n_j)} = M_e(n_i, n_j) \in E^G$; and (c) $\not \exists e_{M_n(n_i),M_n(n_j)} \in E^G \ne M_e(n_i, n_j)$.
	 \end{definition}

	 Note when a graph pattern models a rewrite, $M_n$ and $M_e$ are adjusted accordingly to output the rewrite of a node $n \in N^G$ into its desired value, instead of the set of acceptable nodes from $n \in N^{G_P}$.

	 \begin{definition}[Axiom of equivalence] An axiom $A$ is a semantics-preserving rewrite rule $G' = A(n,G)$ that can arbitrarily modify a program graph $G$, and produces another program graph $G'$ respecting Def.~\ref{def:proggraph} with identical semantics to $G$. We note $A : \langle P_{match}, P_{replace} \rangle$ an axiom, where $P_{match}, P_{replace}$ are graph patterns as per Def.~\ref{def:graphpattern}. The application of axiom $A$ to node $n$ in $G$ is written $A(n,G)$.
	 \end{definition}

	 We can compose axioms to form a complex rewrite sequence.

	 \begin{definition}[Semantics-preserving axiom composition]
	 \label{def:axiomcompos}
	 Given a sequence $S:~ A_1(n_1,A_2(n_2,...,A_m(n_m,G)))$ of $m$ axioms applications. It is a semantics-preserving composition if for each $G_j = A_i(n_i,G_i) \in S$, $P_{match}^{A_i}$ succeeds on the subgraph with root $n_i$ in $G_i$, and $G_j$ is obtained by applying $P_{replace}^{A_i}$ to $n_i$.
	 \end{definition}

	 \begin{theorem}[Program graph equivalence]
	 \label{th:progequiv}
	 Given a program $G$. If $G' = S(G)$ such that $S$ is a semantics-preserving sequence as per Def.~\ref{def:axiomcompos}, then $G \equiv G'$, they are equivalent under the axiom system used in $S$.
	 \end{theorem}

	 This is a direct consequence of using only semantics-preserving axioms, each rewrite cannot individually alter the semantics, so such incremental composition does not. It leads to the formal problem we are addressing:

	 \begin{corollary}[Program graphs equivalence matching]
	 \label{th:progequivmatching}
	 Given two programs $G,G'$. If there exist a semantics-preserving sequence $S$ such that $G' = S(G)$, then $G \equiv G'$.
	 \end{corollary}

	 Note here $=$ means complete structural equivalence between the two graphs: they are identical in structure \emph{and} label/node values. Determining $G = G'$ amounts to visiting both graphs simultaneously e.g. in depth-first search from the root to ensure structural equivalence, and also verifying the same node labels appear in both at the same time. This is trivally implemented in linear time in the graph size.

	 \paragraph*{Language of linear algebra expressions} We have implemented a total of 102 different axioms for our language, made of the multi-type versions of the 13 core restructuring axioms described later in Table~\ref{tab:TransformPct}. They all follow established linear algebra properties. Note different data types have different axioms following typical linear algebra rules, e.g., matrix-multiplication does not commute, but scalar and vector multiplications do. Examples of axioms include $x(yz) \rightarrow (xy)z$, $X-X\rightarrow O$, $-(\vec x - \vec y) \rightarrow \vec y - \vec x$, or $X^{t^t} \rightarrow X$, an exhaustive list is displayed in the Supplementary Material.

	 In our experiments, we presume matrix and vector dimensions are appropriate for the given operation. Such dimension compatibility
	 checks are simple to implement by e.g. introducing additional nodes in the program representation, but are not considered in our test language.




\paragraph*{Examples} We illustrate axiom-based rewrites using axioms presented in later Table~\ref{tab:TransformPct}. Note axiom names follow the structural changes applied. For example, we have $a+b \equiv b+a:~\{a+b\}= Commute(\{+\},\{b+a\})$. $a+b+c \equiv b+c+a:~\{a+b+c\}= Commute(\{+_1\},Commute(\{+_2\},\{b+c+a\})$. Note we refer to different nodes with the same symbol (e.g., $+_2$) subscripting them by their order in a DFS traversal of the program graph, starting from the unique root. We have $0 \equiv a-a:~\{0\}= Cancel(\{-\},\{a-a\})$. These can be combined in complex paths, e.g., $b+c \equiv c+b+(a-a):~\{b+c\}= Commute(\{+\},Noop(\{+\},Cancel(\{-\},\{c+b+(a-a)\})))$. Such axioms are developed for scalars, matrices and vectors, and include complex rewrites such as distributivity rules and transpositions. A total of 102 axioms are used in our system.

\subsection{Space of Equivalences}

We now define the search space being explored in this work, i.e., the exact space of solutions on which the DNN system formally operates, and that we sample for training.

\begin{definition}[Graph of the space of equivalences]
\label{def:graphofequiv}
Given a language $\mathcal{L}$. The directed graph of equivalences between programs is $G^{equiv} = \langle N^{equiv}, E^{equiv}\rangle$ such that $\forall l \in \mathcal{L}, n_l \in N^{equiv}$, and $e^{A_i,x}_{n_i,n_j} : n_i \rightarrow n_j \in E^{equiv}$ iff $n_j \equiv A_i(x,n_i)$, $\forall A_i$ in the axiom system and $x$ a position in $n_i$ where $A_i$ is applicable.
\end{definition}

In other words, the graph has one node per possible program in the language $\mathcal{L}$, and a single axiom application leads to connecting two nodes. We immediately note that $G^{equiv}$ is a (possibly infinite) multigraph, and contains circuits.

\begin{theorem}[Program equivalence with pathfinding]
Given two programs $n_i,n_j \in N^{equiv}$. If there is any path from $n_i$ to $n_j$ in $G^{equiv}$, then $n_i \equiv n_j$.
\end{theorem}

The proof is a direct consequence of Def.~\ref{def:graphofequiv}. In this work, we randomly sample this exact graph to learn how to build paths between arbitrary programs. As it is a multigraph, there will be possibly many different sequences modeled to prove the equivalence between two programs. It is sufficient to expose one to prove equivalence.

\begin{corollary}[Semantics-preserving rewrite sequence]
Any directed path in $G^{equiv}$ is a semantics-preserving rewrite sequence between the programs, described by the sequence of axioms and program position labeling the edges in this path. This sequence forms the proof of equivalence.
\end{corollary}

We believe that ensuring there are possibly (usually) many ways to compute a proof of equivalence in our specific framework is key to enable the DNN approach to learn automatically the pathfinding algorithm for building such proofs. Other more compact representations of this space of equivalences are clearly possible, including by folding nodes in the equivalence graph for structurally-similar programs and folding equivalent paths between nodes. When building e.g. a deterministic algorithm for pathfinding, such space size reduction would bring complexity benefits \cite{kaplan1969regular,barthou2002}. We believe that for the efficient deployment of graph-to-sequence systems, exposing significant redundancy in the space facilitates the learning process. We also alleviate the need to reason on the properties of this space to find an efficient traversal heuristic.

%% file: samplegen.tex

Following the problem formalization in Sec.~\ref{sec:proglangdefs}, the next challenge is to automatically sample the search space graph. The careful design of this step is key: as we let the DNN learn \emph{by example only} what the axioms are  and when they are  applicable, along with what is the general structure of a program, we must carefully sample the space of equivalences to ensure appropriate distributions of the examples. We produce a final dataset of 420,000 tuples $(P1,P2,S)$, a pair of input programs and a possible rewrite sequence between them. 
We outline below its generation principles, extensive details and the algorithms used are presented in Supplementary Material.


\subsection{Random Sample Generation}

Deep learning typically requires large training sets to be effectively deployed, our system is no exception. Hence the need to automate the generation of an arbitrary number of samples. With this process, we can create as large and
varied a dataset as our machine learning approach requires.

We specifically use randomized program generation algorithms that are inspired by a given language grammar. While using a grammar as input is not required, the benefits are immediate in particular for regular LL(1) languages: one can build random parse trees by simply iterating the grammar, randomly choosing between possible productions. The leaves obtained will form a sentence accepted by the language, i.e., a program \cite{bielik16}.

In particular, we skew the pseudo-random generation so that (1) binary operations are more likely to be created than unary operations, and (2) the initial probability that a child of the created graph node will itself be an operation (as opposed to a terminal symbol) is set to 91\%. The algorithm then subtracts a 23\% probability for children at each level of the graph, so that the path length from the root to any leaf does not exceed 6. Note these probabilities and algorithm have been computed to match the size restrictions for our system evaluated in Sec.~\ref{sec:expresults}: handle programs with 30 nodes maximum, and sequences made of 5 axiom applications maximum.

We produce equivalent program samples by iterating pseudo-randomly the axioms on one randomly generated program to produce a rewrite sequence and the associated equivalent program. The process iterates through all nodes of a program graph, and at each node checks which axiom(s) can be applied. E.g., the $+_m$ operator can have the Commute axiom applied, or depending on subtrees it may be allowed to have the Factorleft axiom applied, as discussed in Sec.~\ref{sec:expresults}. Generally we choose to apply or not an operator with 50\% probability, so that \texttt{pe-graph2seq} is forced to rely on analysis of the two programs to determine whether an operator is applied instead of learning a bias due to the local node features.


\subsection{Final Experimental Dataset}
After these generation algorithms are run, a final data preparation process is done to prune the dataset for the learning phase.  Any lexically equivalent program pair (if any) is removed. Importantly, we remove some cases with only 1 or 2 axioms being used once, to slightly bias the dataset to longer rewrite sequences. We also ensured a reasonable statistical distribution of the use of the various axioms.

\begin{table}[h!tb]
\vspace{-.1cm}
{\small
{\begin{tabular}{ |p{2.4cm}|p{3.4cm}|p{1.4cm}| }
 \hline
 Rewrite Rule & Example(s) & Samples using rule \\
 \hline
 Cancel & (A - A) $\rightarrow$ O, (b/b) $\rightarrow$ 1 & 13.0\% \\
 \hline
 Noop & (v - o) $\rightarrow$ v & 29.2\% \\
 \hline
 Double & $A^{t^t} \rightarrow A$, 1/1/x $\rightarrow$ x & 7.5\% \\
 \hline
 Commute & (a + b) $\rightarrow$ (b + a) & 29.5\% \\
 \hline
 DistributeLeft & (a + b)c $\rightarrow$ ac + bc & 28.0\% \\
 \hline
 DistributeRight & a(b + c) $\rightarrow$ ab + ac & 19.6\% \\
 \hline
 FactorLeft & ab + ac $\rightarrow$ a(b+c) & 2.1\% \\
 \hline
 FactorRight & ac + bc $\rightarrow$ (a+b)c & 3.1\% \\
 \hline
 AssociativeLeft & a(bc) $\rightarrow$ (ab)c & 16.6\% \\
 \hline
 AssociativeRight & (ab)c $\rightarrow$ a(bc) & 16.2\% \\
 \hline
 FlipLeft & -(v - w) $\rightarrow$ w-v & 9.7\% \\
 \hline
 FlipRight & a/(b/c) $\rightarrow$ a(c/b) & 23.2\% \\
 \hline
 Transpose & $(AB)^{t} \rightarrow B^{t}A^{t}$,  & 10.1\% \\
 \hline
\end{tabular}}
}
\caption{Distribution of the set of 13 rewrite rule types in the final dataset. The totals add to more than 100\% since a single program pair can require multiple rewrite rules for equivalence proof. In total, 102 axioms are used, when considering different data types and operators.}
\label{tab:TransformPct}
\vspace{-.8cm}
\end{table}

Table~\ref{tab:TransformPct} details the distribution of rewrite rules in the dataset we created, we categorized the axioms by the structural graph changes they implement.
Note specifically for our experiments in Sec.~\ref{sec:expresults}, as we target program graphs made of 30 nodes maximum and sequences using a maximum of 5 axioms applications, we prune from the set any entry that does not fit these restrictions. The full dataset is then split into training, validation and test sets is discussed in Sec.~\ref{sec:expresults:setup}.

%% file: progequivdnn.tex

Prior work explored using graph neural networks (GNNs \cite{li16}) to find
a program embedding usable for machine learning, e.g., \cite{AllamanisICLR18},
GNNs for binary code equivalence checking, e.g., \cite{Xu17}, as well as
using a graph-to-sequence model with attention to analyze and generate human
language, e.g., \cite{Beck18}. But to the best of our knowledge, our work is the first to use a graph-to-sequence
approach to generate a verifiable rewrite rule sequence which proves two program graphs are equivalent.
In this section we discuss the implementation details of these components.

\subsection{\texttt{pe-graph2seq} Deep Neural Network}

Fig.~\ref{fig:Full} overviews the entire system architecture including sample generation, the \texttt{pe-graph2seq} network, and the rewrite checker.
Key design decisions are presented below.

\paragraph*{System components}
The system in Fig.~\ref{fig:Full}  is composed of the following blocks. \textsf{Node initialization} is the process in which the program graph is used to initialize the data structures used by the neural network with correct values, it is a direct procedure which sets up the network.

\textsf{Graph neural network} refers to a neural network that has weights which allow it to learn interrelations between network nodes based on edge connections for the problem set it is trained on.

\textsf{Global attention} \cite{luong15} when used with a graph neural network allows the decoder to pay attention to certain nodes in the graph as it creates each token in the output sequence. For example, a node associated with scalar multiply might get extra attention when deciding that the axiom to apply is commutation.

\textsf{Token embedding} is a neural network layer in which tokens are assigned a learnable multidimensional embedding vector \cite{Mikolov13} which can then be processed by other neural network components.

\textsf{LSTM 2 layers} is referring to 2 layers of Long Short Term Memory (LSTM) neurons, each layer can have hundreds of neurons. An LSTM has 'long' memory in the sense that weights which define its behavior are learned from the training data so it has a long memory regarding all the training data it has seen. It has a 'short' memory in the sense that it is a recurrent neural network unit which can change state as the network processes output tokens. As such, a given LSTM cell could change state when the Commute token is output so that the Commute axiom is not repeated.

\textsf{Token generator} is the final output portion of the network. It learns to output the correct token based on the current LSTM hidden states and the global attention from the graph neural network. As each token is output, it feeds back into the LSTM layer through the embedding layer to affect the next state of the LSTM.

\paragraph*{Graph neural network internal representation}
The sample generation discussed in section \ref{sec:genexamples}
provides input to the \textsf{Node Initialization} module in
Fig.~\ref{fig:Full} to create the initial state of our graph neural
network. For each node in the program graph, a node will be initialized in our
graph neural network. Each node has a hidden state represented by a
vector of floating point values which are used to create an
embedding for the full meaning of the given node. Initially all of the
dimensions of the hidden states of the nodes are set to zero except for 2. 
Given $N$ tokens in our input program language, one of the dimensions from 
1 through $N$ of a node will be set based on the token at the program position 
that the node represents. For example, if the scalar variable $a$ is assigned to
be token 3 in our language, then the $a$ node in Fig.~\ref{treeexamples:1}
would be initialized to 1.0. This is a one-hot encoding similar to that used
in neural machine translation models \cite{HuntDownCitation}. The second non-zero
dimension in our node initialization indicates the tree depth, with the root for the program being at depth 1. We set the dimension $N$+$depth$ to 1.0; hence, the $a$ node in Fig~\ref{treeexamples:1}, which is at level 2 in the graph, would set dimension $N+2$ to 1.
In addition to nodes correlating to all tokens in both input programs, we initialize
a root node for program comparison which has edges connecting to the root nodes of both programs.  The root node does not represent a token from the language, but it is initialized with a 1.0 in a hidden state dimension reserved for its identification.

For a graph neural network, the edge connections between nodes are a
crucial part of the setup. In particular, to match the formulation of our problem, we must ease the ability of the network to walk the input program graphs. We therefore designed a unified graph input, where both program graphs are unified in a single graph using a single connecting root node; and where additional edges are inserted to make the graph fully walkable.

\begin{figure}[h!tb]
\vspace{-.5cm}
\centering
\includegraphics[width=0.5\textwidth]{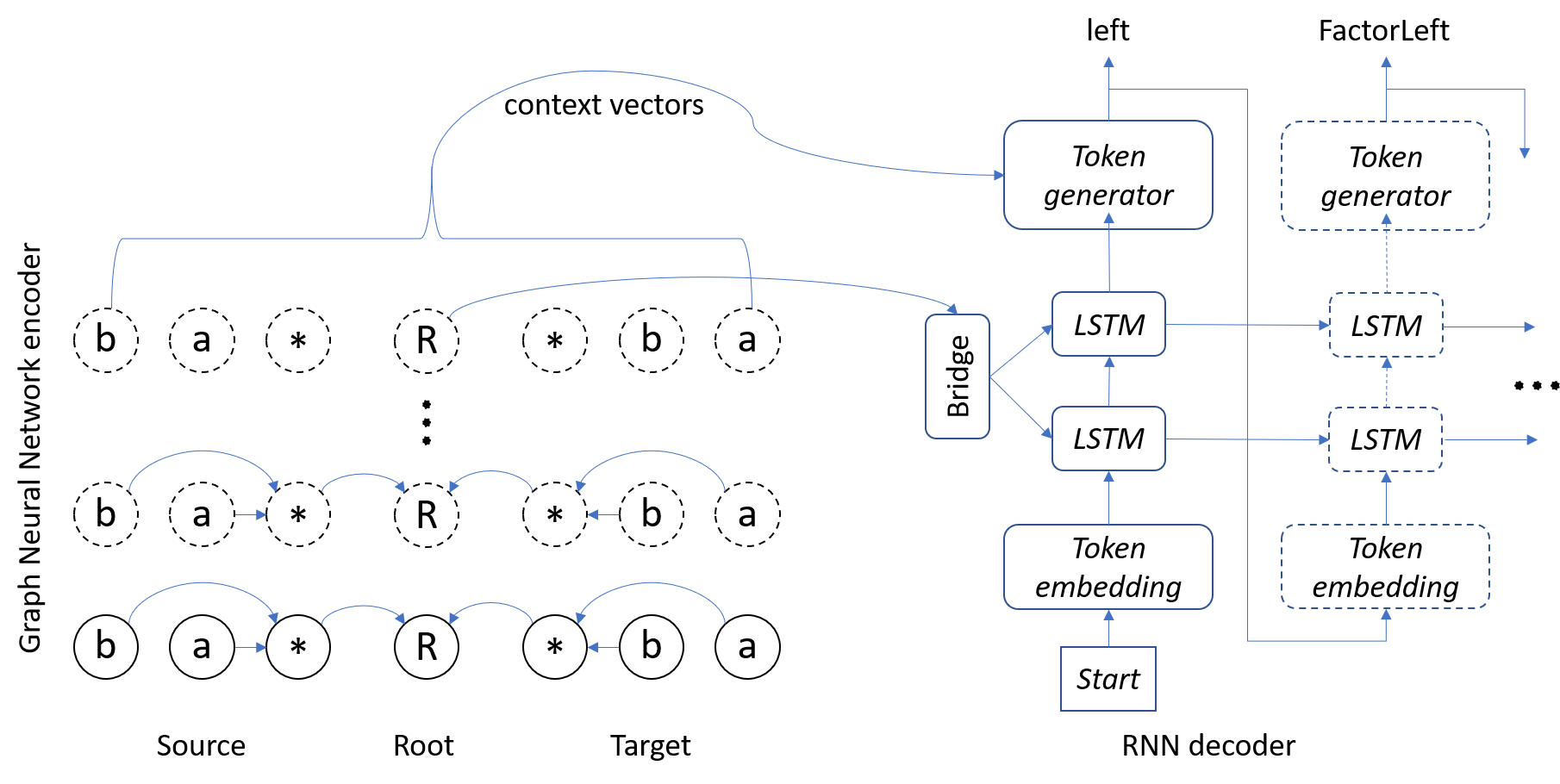}
\caption{Graph-to-sequence neural network data flow details.}
\label{fig:Network}
\vspace{-.5cm}
\end{figure}

In our full model, we support 9 edge types and their reverse edges. The edge types are: 1) left child of binary
op, 2) right child of binary op, 3) child of unary op, 4) root node to
program 1, 5) root node to program 2, 6-9) there are 4 edge
types for the four node grandchilden (LL, LR, RL, RR). After the node
hidden states and edge adjacency matrix are initialized, the network is
ready to begin processing. This initial state is indicated in
figure \ref{fig:Network} by the solid circles in the lower left of the
diagram.

The combination of the root node type and the edges connecting
it to programs 1 and 2 allow the network to learn weights which allow
the graph neural network to 'walk' information from the graph of program
1 through the root node to the graph of program 2 as it creates the node
embeddings necessary for rewrite rule generation. This is a novel feature
of our network not used in prior work with GNNs on program analysis
\cite{AllamanisICLR18,Xu17}.

\paragraph*{Graph neural network processing}
After initialization, the graph neural network iterates in order to
convert the initial node state into the embeddings needed for
rewrite rule generation. For our problem size, we iterate the GNN 10 times.
This process is shown in figure \ref{fig:Network} with the
dotted circles starting with the initial state on the bottom left and
rising to the final state before input to the LSTM-based
decoder. Given an initial hidden state for node $n$ of $x_n(0)$,
$x_n(t+1)$ is computed with a learnable function $f$ which combines
the current hidden state
$x_n(0)$, the edge types $l_{in[n]}$ of edges entering node $n$, the
edge types $l_{out[n]}$ of edges exiting node $n$, and the hidden
states $x_{ne[n]}$ of the neighbors of node $n$:

\[ x_n(t+1) = f(x_n(t),l_{in[n]},x_{ne[n]}(t),l_{out[n]}) \]

Each of the edge types has a different weight matrix for learning,
allowing aggregation of information into a given node related to its
position and function in the full graph of the program. The root node
initial state along with the special edge types connecting it to the
graph trees of the programs are able to learn specific information
regarding rewrite rules as demonstrated by our experimental results.

\paragraph*{Graph neural network output to decoder} Fig.~\ref{fig:Network} shows two ways that the final node values for
the graph are used by the decoder to create the rewrite rules.
First, the final root node value $x_{root}(10)$ is fed through
a learnable bridge function to initialize the 2 layer LSTM of the
decoder network. In this way, the aggregated information of the 2
programs seeds the generation of the rewrite rules. The LSTM layer
updates as each output token $y_{j}$ is generated with a learnable function
based on the current decoder hidden state $h_{j}^d$ at decoder step
$j$ and the previous output token $y_{j-1}$ \cite{Chen19}. Second,
all nodes in the graph can be used by a learnable attention layer
\cite{Bahdanau14}. The attention layer creates a context vector
$c_j$ which can be used by a learnable function $g$ when computing
the probability for generating the $j$th output token $P(y_{j})$:

\begin{equation}
\label{eq:attntokenprob}
P(y_{j} \mid y_{j-1},y_{j-2},...,y_{0},c_j) = g(h_{j}^d, y_{j-1}, c_j)
\end{equation}

By using the root node only for seeding the initial hidden state
$h_{0}^d$ of the decoder, that node and the weights associated
with the connections to the program graphs for programs 1 and 2
are configured so that they learn the information necessary for
starting off the rewrite rule sequences. In parallel, after
the graph neural network iterations complete, the final
embedding for all the nodes in the graphs for programs 1 and 2
are only used by the attention network, so their final
embedding learns to provide useful information during the
rewrite rule generation (i.e., after initialization of the
decoder).




\paragraph*{Beam search}
A typical approach when using sequence-to-sequence systems is to
enable \emph{beam search}, the process of asking for multiple answers to
the same question to the network. It is particularly relevant when
creating outputs which can be automatically
checked \cite{Chen19,ahmed18}. Beam search can be viewed
 as proposing multiple possible paths/rewrite sequences. Given
the stochastic nature of generation model, a beam width of $n$ can
be thought of as creating the $n$ most likely sequences given
the training data the model as learned on. Each proposal
can be checked for validity, the first valid one is outputted by the
system, demonstrating equivalence.
If no sequence is valid, the system
answers the programs are likely not equivalent.
We evaluate in
Sec.~\ref{sec:expresults} beam sizes ranging from 1 to 10, 
showing higher success with larger beams.


%% file: expresults.tex

We now present extensive experimental results, and compare the quality of several neural network approaches to address the problem of program equivalence. We have proceeded incrementally for fine-tuning the final system design, and report on several of these design points below.

\subsection{Implementation Setup}
\label{sec:expresults:setup}

\paragraph*{System implementation} We developed the entire system presented in the OpenNMT-py system \cite{opennmt}, adding on an available prior implementation of gated graph neural networks \cite{li16}. 
Specifically, we developed a general graph neural network encoder within OpenNMT-py, as well as our program graph initialization procedure.

For our training and evaluation experiments, we use systems with Intel Xeon
3.6GHz CPUs and 6GB GeForce GTX 1060 GPUs. OpenNMT-py supports automatic CUDA acceleration of training and inference for our system.


\paragraph*{Training, validation and test sets creation}
For evaluation of our system, we generate sample programs pairs and rewrite rule sequences as discussed in Sec.~\ref{sec:samplegen}. For the initial evaluations, we generate 100,000 total unique samples, separated into 80,000 cases for training, 10,000 for validation of the model during the training process, and 10,000 cases withheld for testing of the selected model. As every tuple $(P1,P2,S)$ in the main dataset is unique, i.e., there is never twice the same pair of programs with the same rewrite sequence, the test set cannot intersect with the training set.

We ensured numerous stringent properties on our test set. $>$99.5\% of the test set cases use at least one input program that does not appear in the training set. 69.07\% of the test set use a rewrite rule sequence that appears in the training set: we ensured about 30\% of the ground truth rewrite sequences in the test set do not even occur in the training set. This proportion has been selected to ensure we verify the system has learned how to reason on programs (e.g., two different pairs of programs may use the same rewrite rule, e.g. $(a+b,b+a)$ and $(c+d,d+c)$, we verify this generalization is learned); and to verify the system can compute new paths/sequences (showing generalization of the concept of incremental application of axioms).

Note that there are 118,278 unique RW sequences in the training data, so, as an example, if we allowed a beam search size of 118,278 instead of 10, and if the network learned to naively output all 118,278 RW sequences from the training data it would fail on 30.93\% of the test data, well below our 95.5\% score. Clearly the network is adding intelligence to the problem.

For our more complex language evaluations, we maintain 10,000
cases for the validation and test sets, but increase the training cases to as
many as 500,000, which approaches the memory limits of the systems we train on.


\paragraph*{Training procedure and parameters}
Our initial investigations with small models are done with 50,000 epochs and a
batch size of 32. Hence, when training with 80,000 training samples each sample
is trained on 20 times.  Our full language trains 400,000 samples for 250,000
epochs on a batch size of 32, hence again each is trained on 20 times.
During training, the majority of our testing runs the validation set
on the model every 10,000 epochs and saves a model for test data
processing every 50,000 epochs. Although we used a validation set for tracking the evolution of the learning quality, we did not use early stopping criteria.

\paragraph*{Evaluation procedure}
Our evaluation consists of multiple scoring methods. As a model is learning
proper matrix weights during training on the training samples, we track the 
typical current per-token prediction accuracy as the model learns to predict the
correct rewrite rule sequences (optionally including the
\texttt{Not\_equal} token). Similarly, when the validation is evaluated, the
token accuracy for the predicted outputs is reported.

However for our test dataset evaluation, we instead of course report the
accuracy of the model to output a correct rewrite sequence
with beam sizes of 1,2,5, or 10. When testing, a ground truth sequence between the two programs is available. The model may or may not produce a sequence that matches exactly this test sample ground truth sequence. We call it a \emph{match} when the model produced the ground truth, and a \emph{correct} sequence when it is a verified correct proof, even when it does not match the ground truth.




\input{tableresults1}

\subsection{Language Complexity and Performance}

As discussed in later Sec.~\ref{sec:expresults:additionalresults}, we iterated numerous possible designs and approaches to figure out the best-working system for this network. In particular, we evaluated simpler approaches before reaching the complexity of our fina design, to ensure a more complex approach was needed.

Table~\ref{tab:ResultLang} shows the result of 12 different experiments and designs. In particular, we incrementally increase the problem complexity from rows 1 to 10, increasing the number of \textsf{Operators} that can be used in any input program, of \textsf{Axioms} used in the rewrite sequence, of \textsf{Operands} in any input program, of the maximal number of nodes in an input program graph (the \textsf{Program length}, directly influencing the size of the graph network), and  the \textsf{Rewrite rule length}, which contains the description of paths from the root node to reach the position of application of an axiom, this is directly related to the maximal graph height, itself determined by the maximal program size. Details on each row are provided in Supplementary Material.

We specifically compare against a sequence-to-sequence (S2S) approach, to quantify the gains brought by employing graph-to-sequence (G2S). When the space is small enough, S2S still performs well, especially using aggressive beam search. We recall that by design of our system testing the correctness of one sequence is trivial and deterministic, so one can easily use large beam sizes without any correctness impact nor major performance penalty during inference. For example, inference of beam 1 is about 15ms for our most complex networks, but beam 10 only takes 16ms. Checking correctness is $<<$ 1ms.

Contrasting rows 2 and 3 displays the merits of the G2S approach for our problem: on this simple problem, in fact G2S gets near-perfect accuracy already. Progressively increasing the complexity of the search space, till row 9 and 10, displays a slow but steady decrease in quality, while still maintaining excellent scores near or above 95\% with beam 10. To reassess the limits of a sequence-to-sequence approach, row 9 and 11 can be contrasted: they operate on the same search space, but S2S peaks at 81\% accuracy, while G2S reaches 95\%.

Row 10 displays the result when learning using also samples of non-equivalent programs, using the ``empty path'' symbol Not\_equal. We evaluated this system to measure the impact of training on only equivalent programs vs. also sampling pairs of unconnected nodes in the equivalences graph. We recall that by design, if no rewrite rule produced is verified as correct, our system  outputs the programs are not equivalent. In other words, whichever the sequence(s) produced by the network, if the two input programs are non-equivalent, the system will \emph{always} output they are not equivalent: no equivalence sequence produced can be verified as correct. So training on only equivalent programs is clearly sensible for such system; furthermore as shown in row 10 vs. 9, even increasing the training set size, training using non-equivalent programs seem to lower the performance slightly.

Our best result (golden model) with the full language has 9545/10000 exact
matches with beam width 10, and 9623/10000 correct
proofs of equivalence (i.e., 78 of the 455 cases without an exact match
still have a legal rewrite rule sequence produced). 

\paragraph*{Manual verifications} We conducted a series of manual verifications of the system used to produce all the above results. First, we are happy to confirm that most likely $AB \ne BA$ given no verifiable equivalence sequence was produced, but that provably $ab=ba$ indeed. We also verified that $A^{t^t}(B+C-C) = AB$, and that $AB\vec v -AB\vec w=AB(\vec v-\vec w)$ which would be a much faster implementation. The system correctly suggests that $AB\vec v -BA\vec w\ne AB(\vec v-\vec w)$. We ensured that $A^t(AA^t)^{-1}A\ne A^t(AA^{-1})^tA$, from a typo we once made when typing the computation of an orthonormal sub-space. We also verified that indeed $AB + AC + aD - aD = A(B+C)$.

In essence, the network has learned each axiom, its valid applicability criteria, and how to sequence them to form a complex rewrite; being trained only from simple pairs of random programs and a sequence describing a labeled path between them in the equivalence space. It learned a generalization of programs, and in particular how to associate tokens for operators to specific axioms whichever their position in the input program. It has also learned how to find a short path in the equivalence graph to build a valid rewrite sequence between the two input programs provided, probably the hardest task of all.

\subsection{Additional Results}
\label{sec:expresults:additionalresults}

In order to design the system, we explored parts of the design space quickly and performed several single training run comparisons between 2 options.
Numerous results are reported in Suppl. material~\ref{sec:suppl:additionalresults}, in Table~\ref{tab:ParamSearch}. They were influential on our final system design. In many cases one model was clearly better than the alternative, driving our design choices.

\paragraph*{Testing simpler models}
In addition to the sequence-to-sequence and graph-to-sequence models, we
explored a feed-forward equal/not equal classifier on a simple version of
our language. That model uses an autoencoder on the program to find an
embedding of the program and then a classifier based on the program embeddings
found. It achieves a 73\% accuracy on the test data, which,
as expected, is much lower than the accuracy rates of 92.4\% with
a graph-to-sequence based classifier on our full language. It also does not produce any verifiable output, contrary to our system.

\begin{figure}[h!tb]
\vspace{-.1cm}
\centering
\includegraphics[width=0.45\textwidth]{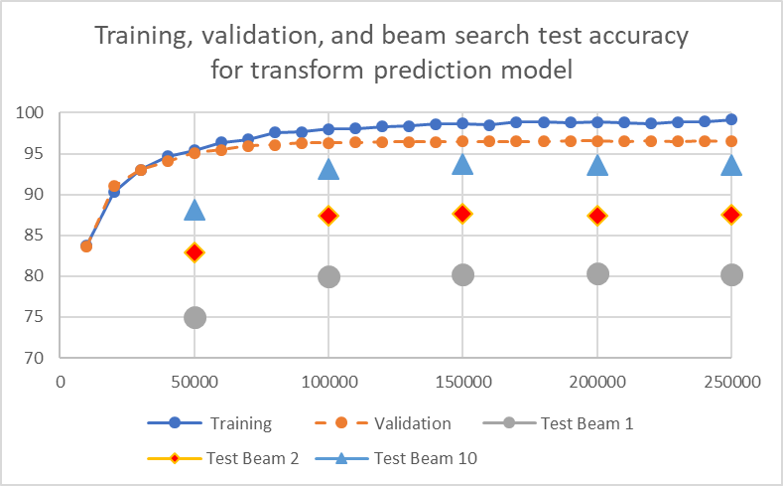}
\caption{Results for network training for rewrite rule generation on 250,000 training samples. The training and validation accuracy are per token of the rewrite sequence; the test accuracy is the score for accurately generating the full sequence with various beam widths.}
\label{fig:ResultAllEq}
\vspace{-.1cm}
\end{figure}

\paragraph*{Evolution of learning quality} Fig.~\ref{fig:ResultAllEq} is for a model trained on 250,000 samples from our full language to generate rewrite rule sequences. The highest test accuracy on beam width 10 is 93.78\%, with the model from iteration 150K. As shown in the figure, the training accuracy continues to increase even as the validation and test accuracies plateau. In order to address this, our final model trains on 400,000 training samples (near the disk space limit available for our testing). That model achieved a test accuracy of 95.45\%.


\subsection{Extensions and Discussions}
\label{sec:ruminations-on-extensions}



\paragraph*{Complex control-flow graph} Given the apparent robustness of our approach to increasingly complex search spaces, we conducted a preliminary study with a node that introduces a cycle in the graph: a loop node (i.e., a back-edge from a node to a leaf).  In Table \ref{tab:ResultLang}, row 12 displays the results. The 2 new operators are used in Fig.~\ref{fig:LoopAST} in Suppl. material. The 'DoX' operator will execute the subgraph some number of times X. The 'DoHalf' operator will execute the subgraph half of X times. Such nodes would model recursive domain decomposition for example, however we restrained from inserting the concept of loop and loop iterators in the language.


The new operator results in 2 new edges in our graph representation (along with 2 new back-edges): there is a 'loopbody' edge type from the loop operator node to the start of the subgraph, and there is a 'loopfeedback' edge type from the variable which is written to each loop iteration. These 2 edge types are shown in the figure. The new $Dohalf$ axiom intuitively states that $DoX(g(y)) = DoHalf(g(g(y)))$ (where $y$ is the variable reused each iteration), and $Dox$ states the reverse. In the results for ID12, 1,412 of the 10,000 test cases involve a loop axiom in the ground truth rewrite rules. Of those cases, 1,351 are matched by the trained network (95.7\%). Additional developments and experiments are needed to ensure we can efficiently manipulate large, complex control-flow graphs, but these results are particularly encouraging.



\paragraph*{Discussions} We have specifically designed our system and its evaluation on a rich language that captures structural changes and properties of a large variety of problems, by modeling input programs as a specific dataflow-like graph. Our results suggest the applicability of this approach to a wide range of problems that can be modeled as finding an (axiom-based) rewrite sequence from one graph to another, a general problem of which program equivalence is only an instance. We evaluated complex graphs including ones containing cycles, however we recall we limited the input graphs to 30 nodes maximum, for the system to complete training in reasonable time. Increasing massively the input program size, e.g., to thousands of nodes, would require larger graph networks to accommodate them, and this puts high stress on the scalability of the training procedure.

Progresses in deep learning frameworks implementation, and/or using other popular systems such as TensorFlow+XLA \cite{abadi2016tensorflow} could possibly significantly accelerate the training time of our experiments and allow the handling of larger problems. Note we did not do any specific effort to optimize the training time in our system.

%% file: tableresults1.tex
\begin{table*}[h]
{\begin{tabular}{ |p{0.4cm}|p{6cm}||p{0.4cm}|p{0.4cm}|p{0.4cm}|p{0.7cm}|p{0.7cm}||p{0.7cm}|p{1cm}||p{0.7cm}|p{0.7cm}| }
 \hline
 ID & Description & \rotatebox{90}{\# Operators} & \rotatebox{90}{\# Axioms} & \rotatebox{90}{\# Operands} & \rotatebox{90}{Program length} & \rotatebox{90}{Rewrite rules length} & \rotatebox{90}{\parbox{2.9cm}{Graph2seq (G2S) \\ or seq2seq (S2S)}} & \rotatebox{90}{Training set size} & \rotatebox{90}{\parbox{2.9cm}{Percent matching \\ with beam width 1}} & \rotatebox{90}{\parbox{2.9cm}{Percent matching \\ with beam width 10}} \\
 \hline
 1 & Rewrite sequence is only single Commute, uses sequence-to-sequence model & 2 & 1 & 10 & 3-19 & 1-5 & S2S & 80,000 & 90.0\% & 96.2\% \\
 \hline
 2 & Rewrite sequence is exactly 2 Commutes, uses sequence-to-sequence model & 2 & 1 & 10 & 5-24 & 3-10 & S2S & 80,000 & 80.3\% & 96.5\% \\
 \hline
 3 & Rewrite sequence exactly 2 Commutes & 2 & 1 & 10 & 5-24 & 3-10 & G2S & 80,000 & 98.9\% & 99.8\% \\
 \hline
 4 & Rewrite sequence exactly 3 Commutes & 2 & 1 & 10 & 7-45 & 5-15 & G2S & 80,000 & 91.4\% & 99.0\% \\
 \hline
 5 & Rewrite sequence 1 to 3 Commutes & 2 & 1 & 10 & 3-45 & 1-15 & G2S & 180,000 & 97.1\% & 99.2\% \\
 \hline
 7 & Commute, Noop, Cancel, Distribute Left, Distribute Right & 4 & 5 & 12 & 3-45 & 1-15 & G2S & 180,000 & 93.1\% & 97.4\% \\
 \hline
 8 & Scalars, Vectors, and Matrices & 16 & 5 & 20 & 3-30 & 1-25 & G2S & 250,000 & 88.3\% & 95.6\% \\
 \hline
 9 & 13 Axioms & 16 & 13 & 20 & 3-30 & 1-25 & G2S & 400,000 & 85.5\% & 95.5\% \\
 \hline
 10 & Rewrite sequence or Not\_equal & 16 & 13 & 20 & 3-30 & 1-25 & G2S & 500,000 & 79.8\% & 93.8\% \\
 \hline
 11 & Test sequence-to-sequence & 16 & 13 & 20 & 3-30 & 1-25 & S2S & 400,000 & 59.8\% & 81.1\% \\
 \hline
 12 & Add loop axioms & 18 & 15 & 20 & 3-30 & 1-25 & G2S & 400,000 & 83.8\% & 94.7\% \\
 \hline
\end{tabular}}
\caption{Description and results for various language complexities studied.}
\label{tab:ResultLang}
\vspace{-.8cm}
\end{table*}

%% file: related.tex



\paragraph*{Theorem provers} The problem of equivalence as we formulated may be solved by other (smart) brute-force approaches, where a problem is solved by pathfinding. This ranges from theorem proving systems like Coq \cite{bertot2013interactive} which supports the formal framework for equivalence we describe in this paper, to (Approximate Probabilistic) Model Checking \cite{clarke1994model,burch1992symbolic,herault2004approximate}, where a program equivalence system can also be built, e.g. \cite{steffen1991data,clarke2003behavioral,visser2003model,namjoshi2000syntactic}. Our contribution is not in the formal definition of program equivalence we presented, semantics-preserving rewrite systems have been studied, e.g. \cite{visser2004program,lucanu2015program,reddy1989rewriting}. But understanding why this particular formalism was well suited to deep learning graph-to-sequence systems was key.
The merits of stochastic search to accelerate such systems has been demonstrated, e.g. \cite{murawski2005probabilistic,herault2004approximate,gogate2012probabilistic}. The novelty of our approach is to develop carefully crafted graph-to-sequence neural networks to automatically learn an efficient pathfinding heuristic for this problem. Our approach is potentially applicable in these areas too, however training scalability can become a challenge if increasing the input representation size excessively. 

\paragraph*{Static program equivalence} Algorithms for static program equivalence have been developed, e.g. \cite{verdoolaege2012equivalence,alias2004recognition,barthou2002,iooss2014program}. These approaches typically restrict to demonstrating the equivalence of different schedules of the operations, possibly dynamically \cite{bao2016polycheck}. In this work we target graph-modifying rewrites (and therefore which alter the operation count). Barthou et 
al. \cite{barthou2002,alias2004recognition} have developed
techniques to recognize algorithm templates in programs. These
approaches are restricted to static/affine transformed programs.
Karfa et al. also designed a method that works for a subset of affine
programs using array data dependence graphs (ADDGs) to represent input
and transforming behaviors.  Operator-level equivalence checking
provides the capability to normalize expressions and establish
matching relations under algebraic transformations
\cite{karfa2013verification}.  Mansky and Gunter used the TRANS
language \cite{kalvala2009program} to represent transformations.  The
correctness proof implemented in the verification framework
\cite{mansky2010framework} is verified by the Isabelle
\cite{isabelle-web} proof assistant.  Other works
also include translation
validation \cite{sorin:proving,necula2000translation}.

\paragraph*{Program analysis with machine learning} Numerous prior work has employed (deep) machine learning for program analysis, e.g. \cite{AllamanisACM18,Alon19,Tufano19,LacomisDIRE2019,Raychev2015,Bavishi17}.
%
code2vec \cite{Alon19} teaches a method for
creating a useful embedding vector that summarizes the semantic
meaning of a snippet of code. Program repair approaches, e.g. \cite{Tufano19,Chen19} are deployed to automatically repair bugs in a program. Output accuracies of up to 20\% on the test set is reported, using sequence-to-sequence models.
Wang et al. \cite{Wang18} learns to extract the rules for Tomita grammars \cite{tomita82} with recurrent neural networks. The learned network weights are processed to create a verifiable deterministic finite automata (DFA) representation of the learned grammar. This work demonstrates that deterministic grammars can be learned with RNNs, which we rely on.

\paragraph*{Graph Neural Networks}
Graph neural networks \cite{Scarselli09,Wu19} use machine learning
to analyze a set of nodes and edges for patterns related to a target problem.
Using a graph-to-sequence network with attention has been analyzed for natural
language processing \cite{Beck18}. Allamanis et al. use graph
neural networks to analyze code sequences and add edge types representing
LastUse, ComputedFrom, and LastWrite to improve the system's ability to
reason about the code \cite{AllamanisICLR18}. Their work achieves 
84\% accuracy on correcting variable misuse cases and provides insights
to useful edge types. 
Structure2vec \cite{Xu17} uses a graph neural network to detect binary code similarity. Structure2vec uses a graph
neural network to learn an embedding from a annotated control flow graph (ACFG)
of a program. This learning process targets the embedding so that equivalent
programs will have equivalent embeddings, reporting precision
scores of 84\% and 85\% on various test datasets for correctly predicting
program equivalence. It only outputs a probability of equivalence, and not a verifiable proof, which is sufficient in their context.


The G2SKGE model \cite{Li19} has a similar graph network structure which uses a node embedding (which they refer to as an information fusion mechanism) in order to predict
relationships between nodes. This technique of using a neural network to understand and predict
node interrelationships is common to our approach.

%% file: conclusion.tex

In this work, we presented \texttt{pe-graph2seq}, the first graph-to-sequence neural network system generating quickly verifiable program equivalence proofs. Evaluated on a rich language for linear algebra expressions, our system  outputs proofs when input programs are equivalent which are verified correct in 96\% of cases. In addition, the system always outputs non-equivalence for non-equivalent programs by design.

We believe the performance of our approach comes in part from using graph neural networks for what they aim to excel at: learning efficient heuristics to quickly find paths in a graph; and the observation that program equivalence can be cast as a path-based solution that is efficiently found by such networks. We demonstrated our approach on a carefully crafted linear algebra language, to expose clearly the various difficulties the system overcame, such as node deletion or subtree manipulation. We believe this has laid the foundations on how to build such deep learning systems for program equivalence in other languages.


%% file: appendix.tex

\nobalance

Figure \ref{fig:Full} overviews the entire system architecture including sample generation, the \texttt{pe-graph2seq} network, and the rewrite checker. In this section we will discuss the implementation details of these components.

\subsection{Generation of Examples}
\label{sec:genexamples}


Machine learning
benefits from large training sets, so in order to produce this data, we
created algorithms that would generate programs meeting a given language
grammar along with target programs which could be reached by applying a
given axiom set. By creating this process, we could create as large and
varied a dataset as our machine learning approach required. 

Algorithm \ref{alg:GenSrc} provides an overview of the full program generation
algorithm. For this generation process, we define a set of operations and
operands on scalars, matrices, and vectors. For our process, we presume matrix
and vector dimensions are appropriate for the given operation as such dimension
checks are simple to implement and are not considered in our procedure. Note the token syntax here is \emph{exactly} the one used by our system:
\begin{itemize}
  \item Scalar operations: \texttt{+s -s *s /s is ns}, where \texttt{is} the unary reciprical and \texttt{ns} is the unary negation.
  \item Matrix operations: \texttt{+m -m *m im nm tm}, where \texttt{im} matrix inversion, \texttt{nm} negates the matrix, and \texttt{tm} is matrix transpose.
  \item Vector operations: \texttt{+v -v *s nv}, where \texttt{nv} is the unary negation.
  \item Scalars: \texttt{a b c d e 0 1}
  \item Matrices: \texttt{A B C D E O I}, where \texttt{O} is the empty matrix and \texttt{I} is the identity matrix.
  \item Vectors: \texttt{v w x y z o}, where \texttt{o} is the empty vector.
\end{itemize}

Initially, \texttt{GenSrc} is called with \texttt{GenSrc("+s -s *s /s
  +s -s *s /s is ns +m -m *m +m -m *m im nm tm +v -v *v +v -v *v
  nv",0.91)}.  In this initial call binary operations are repeated so
that they are more likely to be created than unary operations, and the
initial probability that a child of the created graph node will itself
be an operation (as opposed to a terminal symbol) is set to
91\%. Since the algorithm subtracts a 23\% probability for children at
each level of the graph, at most 6 levels will be created by this algorithm
(i.e., the path length from the root to any leaf does not exceed 6).

Algorithm \ref{alg:GenSrc} starts execution by randomly selecting an
operation from the set provided as input. When \texttt{GenSrc} is called
recursively, the operation set is limited such that the operation produces
the correct type as output (scalar, matrix, or vector). Lines 3 through 15
of the algorithm show an example case where the \texttt{*s} operation is
processed. This operation requires scalar operands. If the probability of
children at this level is met, then \texttt{GenSrc} is called recursively
with only scalar operands available, otherwise a random scalar operand is
chosen. 

The text for algorithm \ref{alg:GenSrc} does not show the process for all 
operations. Certain operations, such as \texttt{*v}, have a variety of 
operand types that can be chosen. The \texttt{*v} operand is a multiplication
which produces a vector. As such, $Av$ (matrix times vector), $bv$ (scalar
times vector), or $vc$ (vector times scalar) are all valid 
options and will be chosen randomly.

\begin{algorithm}[h!tb]
\DontPrintSemicolon
\SetAlgoLined
\KwResult{Prefix notation of computation with parenthesis}
\SetKwInOut{Input}{Input}\SetKwInOut{Output}{Output}
\Input{Ops, P}
\Output{(op L R) or (op L)}
\BlankLine
 
op = select randomly from Ops

// Create subtree for chosen op

\If{op == "*s"}{
    \eIf{random < P}{
        L = GenSrc("+s -s *s /s +s -s *s /s is ns",P-0.23)
    }{
        L = select random scalar operand
    }
    \eIf{random < P}{
        R = GenSrc("+s -s *s /s +s -s *s /s is ns",P-0.23)
    }{
        R = select random scalar operand
    }
    return (op L R)
}
%
%
%
 
\caption{GenSrc}
\label{alg:GenSrc}
\end{algorithm}

After generating a program which follows the grammar rules of our language,
algorithm \ref{alg:GenTgt} will produce a new program along with a set of
rewrite rules which transform the source program to the target program.

Algorithm \ref{alg:GenTgt} receives as input the source program (or
subprogram) along with the \texttt{path} to the current root node of the
source program. If the source program is a terminal symbol, the algorithm
returns with no action taken. Otherwise, the program starts with an
operation and the algorithm proceeds to process options for transforming
the given operation. 

As shown on line 10 of the algorithm, when the operation and children meet the
conditions necessary for a rewrite rule (in this case \texttt{Noop}), the rule
is applied with some probability (in this case 50\%). Note that before 
processing a node, the left and right operands are further analyzed to 
determine their operators and operands as well (or $\bot$ if the child is a
terminal). Processing the left and right operands allows for complex axioms
to be applied, such as distribution or factorization. When a rule is
applied, the rewrite rule is added to the
rewrite rule sequence and 
a new target program is generated for any remaining subtrees. 
When creating the rewrite rules for subtrees, the \texttt{path} varibale is updated as rewrites are done. In the case of \texttt{Noop}, the current node is being updated, so the path is not changed. But in the case of the Commute rule, the return would be generated with \texttt{(op GenTgt(R,path."left ") GenTgt(L,path."right "))} which creates rewrite rules for the prior right and left operands of the \texttt{op} and updates the path used to the new node positions.
With some probability, illegal rewrites can be done; for example, commuting a
subtraction operation or mutating an operation into another. In that case, the
\texttt{GenTgt} process continues to create a target program, but
\texttt{transform\_sequence} is set to \texttt{Not\_equal}.

\begin{algorithm}[h!tb]
\DontPrintSemicolon
\SetAlgoLined
\KwResult{Second program and transform\_sequence}
\SetKwInOut{Input}{Input}\SetKwInOut{Output}{Output}
\Input{ProgA, path}
\Output{ProgB}
\BlankLine

\If{terminal symbol}{return ProgA}

op = find operator of ProgA

L = find left operand of ProgA

R = find right operand of ProgA

Lop,LL,LR = operator and operands of left child

Rop,RL,RR = operator and operands of right child
 
// Randomly apply transform if allowed

\If{random < 0.5 and ((op == "+v" and (L == "o" or R == "o")) or (op == "-v" and R == "o"))}{
    append path."Noop " to transform\_sequence

    // Eliminate unnecessary operator and 0 vector 

    \eIf{L == "o"}{
        return GenTgt(R,path)
    }{
        return GenTgt(L,path)
    }
}
 
\caption{GenTgt}
\label{alg:GenTgt}
\end{algorithm}

After these generation algorithms are run, a final data preparation
process is done which prunes the data set for the learning
algorithm. The pruning used on our final data set insures that the
source and target program pair total to 60 tokens or fewer (where a
token is an operation or terminal), that the graph is such that every node is reachable from the root with a path of length 5 or less,
that there are 5 or fewer rewrite rules applied and that the
rewrite rule token list is 25 or fewer (including left/right
identifiers for location). Also, the pruning insures that there are no
lexically equivalent programs in the process and removes some of the 1
and 2 rewrite rule cases to bias the dataset to longer rewrite
sequences. Table \ref{tab:TransformPct} details the distribution of
rewrite rules created by the full process. Section \ref{sec:Axioms} details
all axioms when variable types and operators are considered.



\subsection{Rewrite checking}
The rewrite checker algorithm is very similar to algorithm \ref{alg:GenTgt}. For
program generation of the target program, algorithm \ref{alg:GenTgt} will check that a node can
legally apply a given rule, apply the rule with some probability, record the action,
and process the remaining program. For rewrite checking, we begin with a
program 1 and a sequence of rewrite rules. We follow the path given by the
rewrite rule sequence, check that a node can legally accept a rule, apply the
rule, and process the remaining rewrite sequence on the adjusted program.
If a rule cannot legally be applied, program 1 is not proven equal to program 2.
If all rules can be legally applied in sequence to program 1, the program is compared
lexically to program 2 and if they match then equivalence has been proven.

\section{Details on Experimental Results}
\label{sec:suppl:additionalresults}

We explore initial language generation using a simple language in order to assess feasibility of different approaches.
For fine tuning network parameters and architectural features, we add more complexity to the language as shown in
table \ref{tab:ResultLang}. Language IDs 1 through 5 are all based on a simple grammar which only allows
the "+" or "-" operators on scalar variables labeled a through j. The only axiom is \texttt{Commute}, which
can be applied on up to 3 nodes in language IDs 4 and 5. The dramatic increase in performance of the
graph neural network for 2 Commute languages is shown by comparing IDs 2 and 3. Language ID 7 adds
the scalar constants 0 and 1, scalar operations * and /, and 4 more axioms. We perform a fair amount of
network development on this model in an effort to maintain high accuracy rates. Language ID 8 expands
the operands to 3 types and hence the number of operators also increase. To account for memory footprint
concerns due growing complexity in our model, we reduce the maximum program size with ID 8. This reduction
also allows us to train larger data sets for more epochs. ID 9 is our full language using our golden model
which we focus on throughout this paper.
ID 10 explores the use model where the model trains to produce a Not\_equal token when the input
programs are not identical. The discussion for table \ref{tab:Proof} will explore the use model in relation
to non-equivalent programs in depth. ID 11 demonstrates on the full model the disadvantage of using
a sequence-to-sequence model for this problem. ID 12 is a forward looking-model which makes a minor
increment to the language to support the analysis of loop rolling and unrolling.

\paragraph*{Exploration of alternate designs}
In order to design the system, we explored parts of the design space quickly and performed several single training run comparisons between 2 options, as shown in Table~\ref{tab:ParamSearch}. 

In cases where 2 options were similar, we
chose the model which ran faster, or run the models a second time to
get a more precise evaluation, or use our experience based on prior
experiments to select an option.

\begin{table}[h!tb]
\small
{\begin{tabular}{ |p{5.2cm}|p{1cm}|p{1.2cm}| }
 \hline
                  & Match  & Match   \\
 Options compared & beam 1 & beam 10 \\
 \hline
 \hline
 1 layer LSTM vs  &  198 & 1380 \\
 2 layer LSTM vs  & 5020 & 9457 \\
 3 layer LSTM     & 4358 & 8728 \\
 \hline
 192 dimension embeddings vs  & 8411 & 9475 \\
 256 dimension embeddings     & 8453 & 9516 \\
 \hline
 256 dimension embeddings vs  & 7033 & 9688 \\
 512 dimension embeddings     & 6905 & 8800 \\
 \hline
 Sequence-to-sequence vs      & 5984 & 8112 \\
 graph-to-sequence            & 8404 & 9488 \\
 \hline
 No edges to grandchild nodes vs      & 9244 & 9728 \\
 Edges to grandchild nodes            & 9284 & 9774 \\
 \hline
 Encoder->Decoder only root node vs      & 8616 & 9472 \\
 Encoder->Decoder avg all nodes          & 7828 & 9292 \\
 \hline
\end{tabular}}
\caption{Example explorations as a single feature or parameter is changed. Each comparison is a distinct experiment, as the entire network and language used was being varied.}
\label{tab:ParamSearch}
\vspace{-.8cm}
\end{table}

Experiments such as these informed our final network architecture. For 
example, in \texttt{pe-graph2seq}, we include 4 edges with learnable weight
matrices from a node to its grandchildren because such edges were found to
improve results on multiple runs.
Li et al. \cite{Li19}  discusses the importance of selecting the optimal process for aggregating
the graph information hence we explore that issue for our network.
Our approach uses the root comparison
node to create aggregate the graph information for the decoder as it performs
better than a node averaging. 
Also clearly shown in these results is the improvement a graph
neural network can provide over the tuned sequence-to-sequence model
provided as part of the OpenNMT system. A sequence-to-sequence model
cannot easily learn the full grammar of the language and the correct nature of
the program as input.

\paragraph*{Including Not\_equal option}

Table \ref{tab:Proof} analyzes the challenge related to a model which only predicts
Equal or Not\_equal for program pairs along with various options which produce
rewrite rules which can be checked for correctness. In all 4 output cases shown,
2 programs are provided as input and programs use our full language model with 16
operators, 13 core axioms (102 total), and 20 operands.

\begin{figure}[h!tb]
{\small
\begin{tabular}{ |p{2.2cm}|p{0.8cm}||p{1.2cm}|p{1.2cm}|p{1cm}| }
 \hline
 Network     &        &           & Predicted & Correct \\
 output      &        & Predicted & Rules     & Rewrite \\
 Description & Actual & NotEq     & or Eq     & Rules   \\
 \hline
 \hline
 Eq or NotEq,    & Eq    & 5.4\%  & 94.6\% & N/A \\
 Beam width 1    & NotEq & 90.4\% & 9.6\% & N/A \\
 \hline
 Rules or NotEq, & Eq    & 6.6\% & 93.4\% & 70.7\% \\
 Beam width 1    & NotEq & 90.9\% & 9.1\% & N/A \\
 \hline
 Rules only,     & Eq    & N/A & 100\% & 87.8\% \\
 Beam width 1    & NotEq & N/A & N/A & N/A \\
 \hline
 Rules only,     & Eq    & N/A & 100\% & 96.2\% \\
 Beam width 10   & NotEq & N/A & N/A & N/A \\
 \hline
\end{tabular}
}
\caption{Table showing alternate options for handling not equal programs\label{tab:Proof}}
\end{figure}

For the first output case, the output
sequence to produce is either \texttt{Equal} or \texttt{Not\_equal}. Given a
false positive rate of 9.6\%, these results
demonstrate the importance of producing a verifiable proof of equivalence
when using machine learning for automated equivalence checking.
For the second output case, the model can produce either \texttt{Not\_equal}
or a rewrite rule sequence which can be checked for correctness. The source
programs for the first and second case are identical: 250,000 equivalent program pairs
and 250,000 non-equivalent program pairs. In the second case, the false positive
rate from the network is 9.1\% (rules predicted for Not\_equal programs), but
the model only produces correct rewrite rules between actual equivalent programs
in 70.7\% of the cases. One challenge with a model that produce rules or
\texttt{Not\_equal} is that beam widths beyond 1 are less usable. Consider that
with a beam width of 1, if the network predicts \texttt{Not\_equal} then the
checker would conclude the programs are not equal (which is
correct for 90.9\% of the actually not equal programs). With a beam width of 10,
there would be more proposed rewrite rules for equal programs to test with, but
if 1 of the 10 proposals is \texttt{Not\_equal}, should the checker conclude they
are not equal? Or should the the checker only consider the most likely prediction
(beam width 1) when checking for non-equivalence? The third and fourth network 
output cases provide an answer. For these 2 cases, the training set is 400,000 
equivalent program pairs - none are non-equivalent. 250,000 of these pairs are
identical to the equivalent programs in the first 2 cases, and 150,000 are new
but were produced using the same random generation process. Note that by requiring
the network to focus only on creating rewrite rules, beam width 1 is able to 
create correct rewrite rules for 87.8\% of the equivalent programs. And now,
since we've remove the confusion of the \texttt{Not\_equal} prediction option,
beam width 10 can be used to produce 10 possible rewrite rule sequences and
in 96.2\% of the cases these rules are correct. Hence, we propose the preferred
use model for pe-graph2seq is to always use the model which is trained for
rule generation with beam width 10 and rely on our rule checker to prevent
false positives. From the 10 rewrite rule proposals, non-equivalent programs
will never have a correct rewrite rule sequence produced, hence we
guarantee there are no false positives. Equivalent programs (within the random
distribution we analyzed) will have a 96.2\% chance of being proven equivalent.

\section{An Example of Back-Edge in the Program Graph}

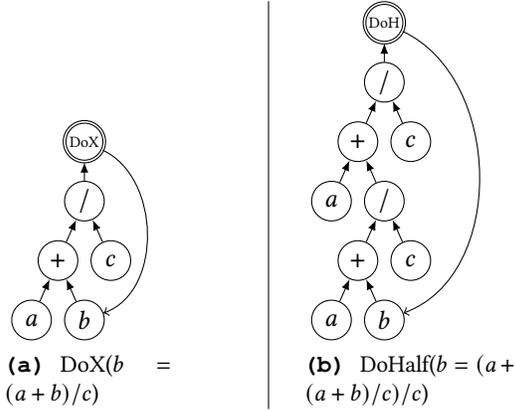
\begin{figure}[h!tb]
\centering
    \begin{minipage}[b]{0.25 \textwidth}
        \centering
        \subfloat [DoX($b=(a+b)/c$)] {
            \begin{tikzpicture} [level distance=2.25em, inner sep=1pt, minimum size=1.5em, sibling distance=2em, edge from parent/.style={draw,latex-}]
                \node [circle, double, draw] (root) {\tiny DoX}
                    child {node [circle, draw] {$/$}
                        child{node [circle, draw] {$+$}
                            child {node [circle, draw] {$a$}}
                            child {node [circle, draw] (leaf) {$b$}}
                        }
                        child {node [circle, draw] {$c$}
                            child{edge from parent[draw=none] node [opacity=0] {}}
                            child{edge from parent[draw=none] node [opacity=0] {}}
                        }
                    };
                \path [->, draw] (root) to [out=335, in=20] (leaf);
            \end{tikzpicture}
        }
    \end{minipage}
    \hfill
    \unskip\ \vrule\
    \begin{minipage}[b]{0.2 \textwidth}
        \centering
        \subfloat [DoHalf($b=(a+(a+b)/c)/c$)] {
            \begin{tikzpicture} [level distance=2.25em, inner sep=1pt, minimum size=1.5em, sibling distance=2em, edge from parent/.style={draw,latex-}]
                \node [circle, double, draw] (root) {\tiny DoH}
                    child {node [circle, draw] {$/$}
                        child{node [circle, draw] {$+$}
                            child {node [circle, draw] {$a$}}
                            child {node [circle, draw] {$/$}
                                child{node [circle, draw] {$+$}
                                    child {node [circle, draw] {$a$}}
                                    child {node [circle, draw] (leaf) {$b$}}
                                }
                                child {node [circle, draw] {$c$}}
                            }
                        }
                        child {node [circle, draw] {$c$}}
                    };
                \path [->, draw] (root) to [out=335, in=20] (leaf);
            \end{tikzpicture}
        }
    \end{minipage}
\caption{Adding loop constructs creates cycles in the program graph.}
\label{fig:LoopAST}
\end{figure}

Figure~\ref{fig:LoopAST} shows an example of DoX and DoHalf.
The new operators result in 2 new edges in our graph representation (along with 2 new back-edges): there is a 'loopbody' edge type from the loop operator node to the start of the subgraph, and there is a 'loopfeedback' edge type from the variable which is written to each loop iteration. These 2 edge types are shown in the figure. The new $Dohalf$ axiom intuitively states that $DoX(g(y)) = DoHalf(g(g(y)))$ (where $y$ is the variable reused each iteration), and $Dox$ states the reverse. 

\section{Full axiom list}
\label{sec:Axioms}

Tables \ref{tab:TransformAll1}, \ref{tab:TransformAll2}, \ref{tab:TransformAll3}, 
and \ref{tab:TransformAll4} show the full 102 axioms supported by our rewrite rules. Many rewrite rules can be applied to all 3 variable types as well as multiple operator types.

{
\begin{figure}[h!tb]
{\small
\begin{tabular}{ |p{2.4cm}|p{1cm}|p{3.4cm}| }
 \hline
 Rewrite Rule & ID & Example(s) \\
 \hline
 Cancel & 1 & (a - a) $\rightarrow$ 0  \\
        & 2 & (b/b) $\rightarrow$ 1  \\
        & 3 & (A - A) $\rightarrow$ O  \\
        & 4 & (v - v) $\rightarrow$ o  \\
 \hline
 Noop  & 5  & (a + 0) $\rightarrow$ a  \\
       & 6  & (0 + a) $\rightarrow$ a  \\
       & 7  & (a - 0) $\rightarrow$ a  \\
       & 8  & (a * 1) $\rightarrow$ a  \\
       & 9  & (1 * a) $\rightarrow$ a  \\
       & 10 & (a / 1) $\rightarrow$ a  \\
       & 11 & (A + O) $\rightarrow$ A  \\
       & 12 & (O + A) $\rightarrow$ A  \\
       & 13 & (A - O) $\rightarrow$ A  \\
       & 14 & (A * I) $\rightarrow$ A  \\
       & 15 & (I * A) $\rightarrow$ A  \\
       & 16 & (v + o) $\rightarrow$ v  \\
       & 17 & (o + v) $\rightarrow$ v  \\
       & 18 & (v - o) $\rightarrow$ v  \\
 \hline
 Double & 19 & -(-a)) $\rightarrow$ a \\
        & 20 & $(a^{-1})^{-1} \rightarrow$ a \\
        & 21 & $-(-A) \rightarrow A$ \\
        & 22 & $(A^{-1})^{-1} \rightarrow A$ \\
        & 23 & $(A^{t})^{t} \rightarrow A$ \\
        & 24 & $-(-v)) \rightarrow v$ \\
 \hline
\end{tabular}
}
\caption{Full axiom count when all type options and other supported permutations are included (part 1 of 4)\label{tab:TransformAll1}}
\end{figure}
}

{
\begin{figure}[h!tb]
{\small
\begin{tabular}{ |p{2.4cm}|p{1cm}|p{3.4cm}| }
 \hline
 Rewrite Rule & ID & Example(s) \\
 \hline
 Commute & 25 & (a + b) $\rightarrow$ (b + a)  \\
         & 26 & (a * b) $\rightarrow$ (b * a)  \\
         & 27 & (A + B) $\rightarrow$ (B + A)  \\
         & 28 & (v + w) $\rightarrow$ (w + v)  \\
         & 29 & (v * a) $\rightarrow$ (a * v)  \\
         & 30 & (a * v) $\rightarrow$ (v * a)  \\
 \hline
 DistributeLeft & 31 & (a + b)c $\rightarrow$ ac + bc  \\
                & 32 & (a - b)c $\rightarrow$ ac - bc  \\
                & 33 & (a + b)/c $\rightarrow$ a/c + b/c  \\
                & 34 & (a - b)/c $\rightarrow$ a/c - b/c  \\
                & 35 & (v + w)*a $\rightarrow$ va + wa  \\
                & 36 & (v - w)*a $\rightarrow$ va - wa  \\
                & 37 & (A + B)C $\rightarrow$ AC + BC  \\
                & 38 & (A - B)C $\rightarrow$ AC - BC  \\
                & 39 & (A + B)v $\rightarrow$ Av + Bv  \\
                & 40 & (A - B)v $\rightarrow$ Av - Bv  \\
                & 41 & (A + B)a $\rightarrow$ Aa + Ba  \\
                & 42 & (A - B)a $\rightarrow$ Aa - Ba  \\
 \hline
 DistributeRight & 43 & a(b + c) $\rightarrow$ ab + ac  \\
                 & 44 & a(b - c) $\rightarrow$ ab - ac  \\
                 & 45 & a(v + w) $\rightarrow$ av + av  \\
                 & 46 & a(v - w) $\rightarrow$ av - av  \\
                 & 47 & A(B + C) $\rightarrow$ AB + AC  \\
                 & 48 & A(B - C) $\rightarrow$ AB - AC  \\
                 & 49 & a(B + C) $\rightarrow$ aB + aC  \\
                 & 50 & a(B - C) $\rightarrow$ aB - aC  \\
 \hline
\end{tabular}}
\caption{Full axiom count when all type options and other supported permutations are included (part 2 of 4)}
\label{tab:TransformAll2}
\end{figure}
}

{
\begin{figure}[h!tb]
{\small
\begin{tabular}{ |p{2.4cm}|p{1cm}|p{3.4cm}| }
 \hline
 Rewrite Rule & ID & Example(s) \\
 \hline
 FactorLeft & 51 & ab + ac $\rightarrow$ a(b+c) \\
            & 52 & ab - ac $\rightarrow$ a(b-c) \\
            & 53 & AB + AC $\rightarrow$ A(B+C) \\
            & 54 & AB - AC $\rightarrow$ A(B-C) \\
            & 55 & Av + Aw $\rightarrow$ A(v+w) \\
            & 56 & Av - Aw $\rightarrow$ A(v-w) \\
            & 57 & Aa + Ab $\rightarrow$ A(a+b) \\
            & 58 & Aa - Ab $\rightarrow$ A(a-b) \\
            & 59 & va + vb $\rightarrow$ v(a+b) \\
            & 60 & va - vb $\rightarrow$ v(a-b) \\
 \hline
 FactorRight & 61 & ac + bc $\rightarrow$ (a+b)c \\
             & 62 & ac - bc $\rightarrow$ (a-b)c \\
             & 63 & a/c + b/c $\rightarrow$ (a+b)/c \\
             & 64 & a/c - b/c $\rightarrow$ (a-b)/c \\
             & 65 & AC + BC $\rightarrow$ (A+B)C \\
             & 66 & AC - BC $\rightarrow$ (A-B)C \\
             & 67 & Av + Bv $\rightarrow$ (A+B)v \\
             & 68 & Av - Bv $\rightarrow$ (A-B)v \\
             & 69 & Aa + Ba $\rightarrow$ (A+B)a \\
             & 70 & Aa - Ba $\rightarrow$ (A-B)a \\
             & 71 & va + wa $\rightarrow$ (v+w)a \\
             & 72 & va - wa $\rightarrow$ (v-w)a \\
 \hline
 AssociativeLeft & 73 & a+(b+c) $\rightarrow$ (a+b)+c  \\
                 & 74 & a(bc) $\rightarrow$ (ab)c  \\
                 & 75 & A+(B+C) $\rightarrow$ (A+B)+C  \\
                 & 76 & A(BC) $\rightarrow$ (AB)C  \\
                 & 77 & A(Ba) $\rightarrow$ (AB)a  \\
                 & 78 & v+(w+x) $\rightarrow$ (v+w)+x  \\
 \hline
\end{tabular}}
\caption{Full axiom count when all type options and other supported permutations are included (part 3 of 4)}
\label{tab:TransformAll3}
\end{figure}
}

{
\begin{figure}[h!tb]
{\small
\begin{tabular}{ |p{2.4cm}|p{1cm}|p{3.4cm}| }
 \hline
 Rewrite Rule & ID & Example(s) \\
 \hline
 AssociativeRight & 79 & (a+b)+c $\rightarrow$ a+(b+c)  \\
                  & 80 & (ab)c $\rightarrow$ a(bc)  \\
                  & 81 & (A+B)+C $\rightarrow$ A+(B+C)  \\
                  & 82 & (AB)C $\rightarrow$ A(BC)  \\
                  & 83 & (AB)a $\rightarrow$ A(Ba)  \\
                  & 84 & (v+w)+x $\rightarrow$ v+(w+x)  \\
 \hline
 FlipLeft & 85 & -(a - b) $\rightarrow$ b-a \\
          & 86 & $(a / b)^{-1} \rightarrow$ b/a \\
          & 87 & $-(A - B) \rightarrow$ (B - A) \\
          & 88 & $-(v - w) \rightarrow$ (w - v) \\
 \hline
 FlipRight & 89 & a/(b/c) $\rightarrow$ a(c/b)  \\
           & 90 & $a/(b^{-1}) \rightarrow$ ab  \\
           & 91 & a-(b-c) $\rightarrow$ a+(c-b)  \\
           & 92 & a-(-b) $\rightarrow$ a+b  \\
           & 93 & A-(B-C) $\rightarrow$ A+(C-B)  \\
           & 94 & A-(-B) $\rightarrow$ A+B  \\
           & 95 & v-(w-x) $\rightarrow$ v+(x-w)  \\
           & 96 & v-(-w) $\rightarrow$ v+w \\
 \hline
 Transpose &  97 & $(AB) \rightarrow (B^{t}A^{t})^t$  \\
           &  98 & $(A+B) \rightarrow (A^{t}+B^{t})^t$  \\
           &  99 & $(A-B) \rightarrow (A^{t}-B^{t})^t$  \\
           & 100 & $(AB)^{t} \rightarrow B^{t}A^{t}$  \\
           & 101 & $(A+B)^{t} \rightarrow A^{t}+B^{t}$  \\
           & 102 & $(A-B)^{t} \rightarrow A^{t}-B^{t}$  \\
 \hline
\end{tabular}}
\caption{Full axiom count when all type options and other supported permutations are included (part 4 of 4)}
\label{tab:TransformAll4}
\end{figure}
}